\documentclass{article}
\pdfoutput=1

\usepackage[preprint,nonatbib]{nips_2018}

\usepackage[utf8]{inputenc} 
\usepackage[T1]{fontenc}    
\usepackage{hyperref}       
\usepackage{url}            
\usepackage{booktabs}       
\usepackage{nicefrac}       
\usepackage{microtype}      
\usepackage{bbm}
\usepackage{color}
\usepackage{graphicx}
\usepackage{caption}
\usepackage{array}
\usepackage{enumitem}
\usepackage{subfigure}
\usepackage{amsmath,amsfonts,amssymb,amsthm}

\DeclareMathOperator{\tr}{tr}
\newcommand{\bB}{\mathcal{B}}

\newcommand{\bm}{\mathbf{m}}
\newcommand{\bS}{\mathbf{S}}
\newcommand{\bk}{\mathbf{k}}
\newcommand{\bmu}{\boldsymbol{\mu}}

\newcommand{\bx}{\mathbf{x}}
\newcommand{\by}{\mathbf{y}}

\newcommand{\bz}{\mathbf{z}}
\newcommand{\bff}{\mathbf{f}}
\newcommand{\bu}{\mathbf{u}}
\newcommand{\bh}{\mathbf{h}}
\newcommand{\bb}{\mathbf{b}}

\newcommand{\btheta}{\boldsymbol{\theta}}

\newcommand{\Op}{\mathcal{O}}

\providecommand{\theoremname}{Theorem}
\providecommand{\propositionname}{Proposition}

\title{Fully Scalable Gaussian Processes using Subspace Inducing Inputs}

\author{
  Aristeidis Panos\\
  Department of Statistical Science\\
  University College London\\
  \href{mailto:aristeidis.panos.15@ucl.ac.uk}{\texttt{aristeidis.panos.15@ucl.ac.uk}}
  \And
  Petros Dellaportas\\
  Department of Statistical Science\\
  University College London,\\
  Department of Statistics\\
  Athens University of Economics and Business,\\
  and The Alan Turing Institute\\
  \href{mailto:p.dellaportas@ucl.ac.uk}{\texttt{p.dellaportas@ucl.ac.uk}}
  \And
  Michalis K. Titsias\\
  Department of Informatics\\
  Athens University of Economics and Business\\
  \href{mailto:mtitsias@aueb.gr}{\texttt{mtitsias@aueb.gr}}
}

\begin{document}

\maketitle

\begin{abstract}
We introduce fully scalable Gaussian processes, an implementation scheme that tackles the problem of treating a high number of training instances together with high dimensional input data.  Our key idea is a representation trick over the inducing variables called subspace inducing inputs.  This is combined with certain matrix-preconditioning based parametrizations of the variational distributions that lead to simplified and numerically stable variational lower bounds.  Our illustrative applications are based on challenging extreme multi-label classification problems with the extra burden of very large number of class labels.  We demonstrate the usefulness of our approach by presenting predictive performances together with low computational times
in datasets with extremely large number of instances and input dimensions.  
\end{abstract}

\section{Introduction \label{sec:introduction}} 

Advances in sparse Gaussian processes (GPs) using inducing variables \cite{csato-opper-02, lawrence-seeger-herbrich-01, seeger03, candela-rasmussen-05, Snelson2006, 
titsias2009variational, hensman2013gaussian, Buietal2017}  and stochastic optimization \cite{Hoffman-2013}  have allowed to 
reduce the initial $\Op(N^3)$ complexity to $\Op(M^3)$ where $M \ll N$ is the number of optimizable 
variational inducing variables in problems with $N$ instances. However, the  $\Op(M^3)$ complexity implies that the input data 
dimensionality, denoted by $D$, is small or of the order of $M$, which does not hold in modern 
machine learning applications such as those arising, for example, in web crawling, gene sequencing and extreme multi-label classification. In such applications the actual time complexity of the most advanced sparse GP methods that 
optimize inducing inputs  is $\Op(D M^2)$.  
This issue has been the major impediment to widespread use of GP's in problems with high dimensional input spaces.

Here, we tackle the problem of treating high dimensional input data in GPs. 
We adopt the sparse GP variational inference framework using inducing variables 
\cite{titsias2009variational} and its stochastic and non-Gaussian likelihood variants \cite{hensman2013gaussian, Lloyd2015,hensman2015scalable, Dezfouli2015, sheth15}.
This implementation strategy is enriched by our key idea which is a representation trick over the inducing inputs called subspace inducing inputs which allows us to drop the computational cost from $\Op(D M^2)$ to  $\Op(R M^2)$, 
where $R \ll D$. This is achieved by precomputing a fixed set of basis vectors in the input space and 
then optimize the inducing inputs by learning linear combinations of these basis vectors. 
A second  generally applicable technique we introduce is that we derive simplified and numerically 
stable variational lower bounds by considering parsimonious parameterizations of the variational distribution 
over the inducing variables with $2 M$ parameters. 
Thus, we derive a robust sparse GP algorithm that can scale to arbitrarily large numbers of data instances and input dimensionality. We call the resulting implementation strategy as {\em fully} scalable Gaussian processes because we achieve scalability in both  $N$ and $D$.  

We illustrate the performance of fully scalable GPs in a a series of  extreme multi-label classification problems with real datasets.  
Multi-label classification is a supervised learning problem where data instances are associated with 
multiple classes \cite{Tsoumakas2007,Read2011, Zhang2013, Gibaja2014, Gibaja2015}. It can be viewed as
a generalization to the more traditional multi-class classification problem
where each data point can belong only to a single class. Multi-label learning has attracted a lot  
of attention in the recent 
literature due to its numerous applications ranging from text and image classification to 
computational advertising and recommender systems \cite{Gibaja2014, Gibaja2015, Prabhu-2014,jain2016extreme}. 
Two main challenges in multi-label learning are: (i) the modelling challenge associated with introducing 
suitable models to capture the correlation across  different labels, and (ii) the computational or scalability 
challenge associated with dealing with datasets having very large number of labels $K$, training instances $N$ and input dimensions $D$. From a GP perspective multi-label learning shares similarities with the standard approaches for multi-task 
 or multi-output Gaussian regression suitable  for real-valued output data \cite{SLFM2005,bonilla2008multi,alvarez2012kernels}. The   
 difference is that in multi-label learning the output data are binary, thus  requiring Bernoulli or binary regression type of likelihoods. Based on this, we  
 introduce a multi-label extension of the semiparametric latent factor model \cite{SLFM2005}  that allows us to capture the correlation of multiple labels using a small set of shared latent 
 GP functions. 
 
As a result, for the multi-label classification problems we present a fully scalable GP model that scales well with $N$, $D$ and  $K$ and it has performance close to the state-of-the-art.  
The key element of the method is that we optimize subspace inducing points using gradient-based methods, that allows us to cope with extremely high dimensional input spaces involving possibly thousands dimensions and we show that such optimization can significantly improve predictive performance and can be close to the corresponding performance 
of optimizing the inducing inputs on their original input space.       

The remainder of the paper has as follows. Section  \ref{sec:theory} gives a brief introduction to variational sparse GPs,  describes the subspace inducing inputs trick for dealing with high dimensional spaces and our parsimonious parameterization of the variational distribution. Section \ref{sec:model} presents our modelling proposal to the multi-label GP model  and how scalable variational inference is achieved. Section \ref{sec:results} demonstrates the method using a series of both small and large scale multi-label datasets and finally the paper concludes with a discussion in Section \ref{sec:discussion}.

\section{Theory  \label{sec:theory}} 

Section \ref{sec:backgroundsparseGPs} provides background on variational 
sparse GPs by noting also that the actual complexity of such methods is 
$\Op(D M^2  + M^3)$ so that the term $\Op(D M^2)$ can dominate for very 
large (hundreds of thousands) or even 
moderate-size $D$. Section \ref{sec:SubspaceInducingInputs} presents
our main methodological contribution of the paper, i.e.\ the introduction    
of subspace inducing inputs that lead to $\Op(R M^2  + M^3)$  time complexity, where $R \ll D$.    
Section \ref{sec:OMparam} presents a novel $\Op(M$) 
parametrization of the variational distribution over the inducing variables 
that can further speed up and numerically stabilize the 
sparse GP training procedure.  

\subsection{Background on variational sparse GPs \label{sec:backgroundsparseGPs}}

Suppose a training dataset $\mathcal{D} = (X, \by)$ where $X = \{ \bx^{(i)} \}_{i=1}^N \in \mathbb{R}^{N \times D}$ is the design matrix of the input vectors and $\by = \{ y^{(i)} \}_{i=1}^N  \in \mathbb{R}^{N}$ the corresponding vector
of outputs where we assume for simplicity that each $y^{(i)}$ is a scalar. Each 
output $y^{(i)}$ is generated through a latent variable $f^{(i)} \equiv f( \bx^{(i)})$ 
so that the full likelihood is $p(\by |\bff) = \prod_{i=1}^N p(y^{(i)} | f^{(i)})$  where  
$\bff$ collects all $f^{(i)}$s. We further assume that $\bff$ follows a GP, i.e.\ 
it is an $N$-size sample of a full random function $f \sim \mathcal{G P}\left( 0, k (\bx^{(i)}, \bx^{(j)}) \right)$, 
so that $p(\bff) = \mathcal{N}(\bff|{\bf 0}, K_X)$ where $K_X$ is the covariance matrix obtained 
by evaluating the kernel function $k (\bx^{(i)}, \bx^{(j)})$ at $X$. The kernel function  
typically depends on hyperparameters $\btheta$ (although for notational simplicity we suppress $\btheta$ throughout).  
The joint distribution over $(\by,\bff)$ is written as $p(\by, \bff) = p(\by|\bff) p(\bff)$ while the marginal likelihood 
takes the form     
$
p(\by) = \int  p(\by|\bff) p(\bff) d \bff. 
$
Even when $p(\by|\bff)$ is Gaussian the computation of the marginal likelihood and 
the predictive posterior process require $\Op(N^2)$ storage and $\Op(N^3)$ time.  
To obtain approximate or sparse solutions several methods are based on inducing variables 
\cite{csato-opper-02, lawrence-seeger-herbrich-01, seeger03, candela-rasmussen-05, Snelson2006, titsias2009variational, hensman2013gaussian, Buietal2017}. Here, we focus on the variational sparse GP 
framework in \cite{titsias2009variational}  
%
which augments the initial joint distribution $p(\by,\bff)$ with 
additional latent function values
%
$\bu \in \mathbb{R}^M$ evaluated at some inputs $Z \in \mathbb{R}^{M \times D}$, so that the 
augmented joint becomes 
\begin{equation}
p(\by, \bff) = p(\by|\bff) p(\bff|\bu) p(\bu) 
\label{eq:joint_sparse_augmented},
\end{equation}
where $p(\bu) =  \mathcal{N} ( \bu | \mathbf{0}, K_Z )$ is the marginal GP prior over $\bu$ and $K_Z$ is the $M \times M$ covariance matrix obtained by evaluating the covariance function at $Z$, while $p(\bff | \bu)$ is the conditional 
GP prior given by
$p(\bff | \bu) =  \mathcal{N} ( \bu | K_{X Z} K_Z^{-1} \bu, K_X -  K_{X Z} K_Z^{-1} K_{Z X} )$,
with $K_{Z X}$ being the cross-covariance matrix between $X$ and $Z$. 
The vector $\bu$ is referred to as inducing variables and $Z$ as the inducing or pseudo inputs \cite{Snelson2006,candela-rasmussen-05}. In the variational sparse GP method \cite{titsias2009variational} 
$Z$ plays the role of a variational parameter that can be optimized 
to improve the approximation. For any value of $Z$ the augmentation in \eqref{eq:joint_sparse_augmented} 
does not change the model 
(i.e.\ the exact marginal likelihood and the posterior process are invariant to the value of $Z$), 
however by applying a certain variational 
approximation in the space of $(\bff,\bu)$ we can both reduce the time complexity and also treat 
$Z$ as a variational parameter that can  be tuned to improve the sparse GP approximation.  
This is achieved by choosing the approximate posterior to be
\begin{equation}\label{eq:approx_posterior}
q(\bff, \bu) = p(\bff | \bu) q(\bu), 
\end{equation}
where $p(\bff | \bu)$ is the conditional GP prior that appears also in the joint \eqref{eq:joint_sparse_augmented},
while $q(\bu) = \mathcal{N}(\bu | \bm, \bS)$ is a Gaussian variational distribution 
over the inducing variables and $(\bm, \bS)$ are variational parameters. 
The work in \cite{titsias2009variational} is concerned with standard GP repression where 
$q(\bu)$ is treated optimally and the resulting collapsed variational lower bound  
(i.e.\ with $q(\bu)$ optimally removed from 
the optimization) is maximized wrt the variational parameters $Z$ and the 
kernel hyperparameters $\btheta$. Each optimization step of the lower bound it  
scales as $\Op(N M^2)$. 

To deal with big data \cite{hensman2013gaussian} extended  
the variational sparse GP method by combining it with stochastic optimization so that the complexity 
per optimization step is reduced from $\Op(N M^2)$ to  $\Op( M^3)$. This can be further combined with   
approximations that deal with non-Gaussian likelihoods \cite{hensman2013gaussian,Lloyd2015,Dezfouli2015,hensman2015scalable}
to obtain a general training procedure that optimizes over $(\bm, \bS, Z, \btheta)$ 
and maximizes the following lower bound on the log marginal likelihood,
\begin{align}
\mathcal{F}(\bm, \bS, Z, \btheta) 
& = \sum_{i=1}^N \mathbb{E}_{q(f^{(i)})} \left[  \log p(y^{(i)} | f^{(i)}) \right] - \text{KL}[q(\bu) || p(\bu)],
\label{eq:lowerbound_sparse}
\end{align}
where $q(f^{(i)}) = \int p( f^{(i)} | \bu) q(\bu) d \bu$. When
maximizing this bound each stochastic gradient ascent step costs $\Op(M^3)$.

However, all previous work on sparse GPs does not take into account the dimensionality of the input space $D$ 
when expressing time complexities and somehow $D$ is assumed to be small or of the order of $M$. 
Since $D$ appears in the lower bound only through the computation of the covariance matrix 
$K_Z$ and the cross covariance matrix $K_{X_b Z}$, where $X_b$ is a minibatch of size $|X_b| \sim \Op(M)$,  
the time complexity with respect to $D$ is clearly $\Op(D M^2)$ since evaluating any standard
kernel function on each pair of instances scales as  $\Op(D)$. Thus, each optimization step of the bound 
in \eqref{eq:lowerbound_sparse} actually scales overall as $\Op(D M^2 + M^3)$ and when $D$ is larger 
than $M$  the term  $\Op(D M^2)$ dominates. For instance, in a dataset as MNIST where $D=784$ and $M=500$ 
the optimization of the bound will roughly be of order $\Op(M^3)$, while in other datasets with even slightly 
larger $D$, such as the CIFAR-10 dataset where $D=3072$, the term $\Op(D M^2)$ dominates and  
thus resulting in slower training. 
Next in order to express fully scalable GPs, 
both in terms of $N$ and $D$, we introduce  
subspace inducing inputs.

\subsection{Subspace inducing inputs \label{sec:SubspaceInducingInputs} } 


Learning 
the inducing inputs $Z$  (as opposed to fixing them to a subset of training instances) is rather crucial in 
order to obtain good approximations, as was initially observed in 
GP regression \cite{titsias2009variational} but also more 
recently in non-Gaussian likelihoods such as in GP classification \cite{hensman2015scalable,HernndezLobato2016ScalableGP}.
By optimizing over $Z$ we are reducing the KL divergence between the approximate 
posterior and the exact posterior process \cite{Matthews16,titsias2009variational} in a way 
that both the likelihood $p(\by|\bff)$ and the kernel function are taken into account.  
The benefit from optimizing $Z$ can be even more profound in high 
dimensions where simple heuristics such as placing $Z$ in a grid or setting it 
using clustering could be non-applicable or sub-optimal. 
However, free-form gradient-based optimization over $Z$ in high dimensions is challenging 
since at each step it requires computing gradients over 
$D M$ parameters, and clearly when $D$ is very large this becomes very expensive. To cope with this, 
we propose to restrict the gradient-based optimization over $Z$ in 
a data-informed lower dimensional manifold or subspace. 
 
Our key idea is to represent $Z$ through the use of a precomputed basis so as the 
optimizable parameters in $Z$ will reduce from $\Op(D M)$  
to $O(R M )$ where $R \ll D$.  
We consider the case of a linear kernel function $k(\bx', \bx)= \bx'{^\top} \bx$
while the popular squared exponential kernel  can be treated similarly  
(see section \ref{sec:sqe} in Appendix).  
Suppose we have precomputed a basis of $R$ vectors stored as  
separate rows in matrix $\widetilde{X} \in \mathbbm{R}^{R \times D}$. For instance, 
 $\widetilde{X}$ can be obtained either by clustering the rows of $X$ or by applying 
a matrix decomposition technique. In all our experiments in Section \ref{sec:results},
we construct $\widetilde{X}$ using singular value decomposition, i.e.\ by computing the $R$ right-singular 
vectors that correspond to the $R$ largest singular values of $X$ using 
the efficient subset singular-value decomposition (SVDs) algorithm \cite{baglama2005augmented}.  
We then parametrize $Z$ as
\begin{equation}\label{eq:AX_tilde}
Z = A \widetilde{X}, 
\end{equation}
where $A \in \mathbbm{R}^{M \times R}$ is a real-valued matrix of tunable/variational parameters. This allows to construct  $Z$ so that each individual inducing input $\bz_i \in \mathbbm{R}^D$ is 
a linear combination of the basis vectors in $\widetilde{X}$ and where the weights in this 
combination are given by the $i$-th row of $A$. At each optimization step of the lower bound 
we need to compute the square kernel matrix $K_Z$ and the cross kernel matrix $K_{X_b Z}$. We 
can compute $K_Z$ as follows     
\begin{equation}
K_Z = Z Z^\top = A \widetilde{X} \widetilde{X}^\top A^\top = A K_{\widetilde{X} } A^\top, 
\label{eq:kZ}
\end{equation}
where crucially the $R \times R $ matrix $K_{\widetilde{X}}$ can be precomputed and stored 
before the optimization starts, i.e.\ while 
such computation requires $\Op(D M^2)$ time it needs to be performed only once.      
Then, any subsequent computation of $K_Z$ and its gradient wrt $A$ 
costs $\Op(R M^2)$. Similarly, the computation of the cross covariance matrix 
$K_{X_b Z}$ reduces to 
\begin{equation}
K_{X_b Z} = X_b Z^\top = (X_b \widetilde{X}^\top) A^\top = K_{X_b \widetilde{X}} A^\top,  
\label{eq:KXbZ}
\end{equation} 
where again the computation of $K_{X_b \widetilde{X}}$ can be done only once beforehand, i.e.\
by precomputing the whole $N \times R$ matrix $K_{X \widetilde{X}}$ and then selecting for any 
minibatch the corresponding block. Note also that for many datasets, 
as the majority of the multi-label classification datasets with extreme input dimensionality, 
$X$ is a sparse matrix and therefore instead of keeping the full matrix 
$K_{X \widetilde{X}}$ in memory, we could alternatively perform the matrix multiplication   
$X_b \widetilde{X}^\top$ at each minibatch optimization step with low computational cost 
by taking advantage of the sparsity of $X_b$.   


The matrix $A$ can be initialized by the $M$ centroids given by k-means with $M$ clusters over the matrix $U S \in \mathbb{R}^{N \times R}$ where $U \in \mathbb{R}^{N \times R}$ contains as columns the left-singular vectors of $X$ and $S \in \mathbb{R}^{R \times R}$ is a diagonal matrix with the $R$ largest singular values of $X$. Notice that both $U$ and $S$ are obtained 
by the singular-value decomposition of $X$, 
 when we construct the basis 
$\widetilde{X}$. 

Section \ref{sec:sqe} in Appendix presents full details about how to apply the 
above technique to the squared exponential kernel, 
while the application to other kernels is left for future work.

\subsection{$\Op(M)$  parametrization of the $q(\bu)$ distribution \label{sec:OMparam}}

Here, we develop a computationally economical and simultaneously flexible 
parametrization of the Gaussian variational distribution 
$q(\bu) = \mathcal{N}(\bu | \bm, \bS)$ in order to speed up the optimization 
of the bound in \eqref{eq:lowerbound_sparse}. Note that a naive 
free-form parametrization of $(\bm, \bS)$, where e.g.\  
$\bS$ is further parametrized based on the Cholesky decomposition, can lead to slow 
convergence due to the strong dependence of $(\bm, \bS)$ with the kernel matrix $K_Z$ 
from the prior $p(\bu)$. To expose such dependence and motivate our method let us re-write the lower bound in 
\eqref{eq:lowerbound_sparse} so that in the first data term instead of marginalizing out 
$\bu$ we marginalize out $\bff$ 
so that 
\begin{align}\label{eq:lowerbound_sparse2}
\mathcal{F}(\bm, \bS, Z, \btheta) 
= \sum_{i=1}^N \mathbb{E}_{q(\bu)} \left[  \log G (y^{(i)}, \bu) \right] - \text{KL}[q(\bu) || p(\bu)],
\end{align}
where $\log G (y^{(i)}, \bu) = \mathbb{E}_{q( f^{(i)} | \bu)} [\log p(y^{(i)}, \bu)]$. A 
straightforward derivation similar to \cite{opper2009variational} can reveal 
that at maximum it holds  
$\bm = K_Z \bmu$ and $\bS = (K_Z^{-1} + \Lambda^{-1})^{-1}$ 
for some vector $\bmu$ and some full (non-diagonal) positive definite matrix $\Lambda$ associated with 
the second derivatives of the first data term in the above bound. This suggests to 
parametrize $q(\bu)$ in terms of $(\bmu, \Lambda)$ in order to take advantage of the 
preconditioning with the kernel matrix $K_Z$. However, this can still lead to slow 
optimization because the full $\Lambda$ matrix requires optimizing over $\Op(M^2)$ 
parameters.  Therefore, here we propose to simplify this parametrization by replacing 
$\Lambda$ with a diagonal covariance matrix $\Sigma$ leading to  the parametrization 
\begin{equation}
\bm = K_Z \bmu, \ \  \bS = (K_Z^{-1} + \Sigma^{-1})^{-1} = K_Z - K_Z (K_Z + \Sigma)^{-1} K_Z, 
\end{equation}  
where $\bmu \in \mathbb{R}^{M}$ is a real-valued vector of tunable variational parameters 
and $\Sigma$ is a diagonal positive definite matrix (i.e.\ with each diagonal element restricted to be non-negative) parametrized by $M$ additional variational parameters. Thus, overall 
$q(\bu)$ is parametrized by $2 M$ 
variational parameters 
while all the remaining structure comes from a careful preconditioning with the model kernel matrix $K_Z$. 
The above parametrization has been used before for full (i.e.\ non-sparse) GPs 
in \cite{opper2009variational,Damianou11} in order to parametrize a full $q(\bff)$, and it was 
motivated by the stationary conditions satisfied by the optimal $q^*(\bff)$ in a full GP  
variational approximation \cite{opper2009variational} where at maximum the  
covariance is $(K_X^{-1} + \Sigma^{-1})^{-1}$ with $\Sigma$ being a diagonal 
positive definite matrix. In our sparse GP setting the $q(\bff)$ 
induced by the above choice of $q(\bu)$ 
is $q(\bff) = \mathcal{N}(\bff | \bm^f, \bS^f)$ where
\begin{equation}
\bm^f  = K_X \bmu, \ \ \bS^f = K_X - K_{X Z} (K_Z + \Sigma)^{-1} K_{Z X},
\label{eq:muSf}
\end{equation}
which can recover the optimal $q^*(\bff)$ when we place the inducing inputs on the
training inputs, i.e.\ when $Z=X$. In other cases the restricted covariance in $q(\bff)$ will 
not be able to match exactly the optimal one of $q^*(\bff)$, but still in practice it tends to be very 
flexible especially when we optimize over the inducing inputs $Z$ so that a posteriori $\bff$ is 
well reconstructed by $\bu$. 

Furthermore, the above parametrization of $q(\bu)$  leads to a numerically stable and 
simplified form of the lower bound. Specifically, the KL divergence 
term in \eqref{eq:lowerbound_sparse} reduces to 
\begin{equation}
\text{KL} [q(\bu) || p(\bu)] = \frac{1}{2} \bmu^\top K_Z \bmu - 
\frac{1}{2} \tr \left(  (K_Z + \Sigma)^{-1} K_Z \right) + \frac{1}{2} \log |K_Z + \Sigma| - \frac{1}{2} \log |\Sigma|, 
\end{equation}
while each marginal $q(f^{(i)})$ in the expectations of the first data term in \eqref{eq:lowerbound_sparse} 
becomes $q(f^{(i)}) = \mathcal{N}( f^{(i)} | m^{(i)}, s^{(i)})$ 
where $m^{(i)}$ and $s^{(i)}$ are the $i$-th elements of the vectors $\bm^f$ and $\bS^f$ in \eqref{eq:muSf}. 
Therefore, the overall bound in \eqref{eq:lowerbound_sparse} obtains a quite simplified 
and numerically stable form because of the cancellation of all 
inverses and determinants of $K_Z$. At each optimization the only 
matrix we need to decompose using Cholesky is $K_Z + \Sigma$, which is in an already 
numerically stable form due to the inflation of the diagonal of $K_Z$ with $\Sigma$. 
In a practical implementation we can constrain the diagonal variational parameters of $\Sigma$ 
to be larger than a small value (typically $10^{-6}$) to ensure numerical stability throughout optimization.

\section{Application to multi-label classification \label{sec:model}} 

Here, we apply the fully scalable GP framework to multi-label 
classification. In this problem the training dataset $\mathcal{D} = (\bx^{(i)}, \by^{(i)})_{i=1}^N$ is such that  
each output $\by^{(i)} \in \{-1,1\}^K$ is a binary vector that indicates 
the class labels assigned to $\bx^{(i)}$, so that $y_k=1$ indicates  presence of the 
$k$-th label while $y_k=-1$ indicates absence. 
We will collectively denote all binary labels by 
$Y \in \{ -1 , 1 \}^{N \times K}$ so that rows of these matrices store respective data points. 
As a suitable GP-based probabilistic model for these data we consider a multi-label extension of the  
semiparametric latent factor model (SLFM) of \cite{SLFM2005} that combines a linear latent variable model 
with GPs. Specifically, SLFM is a general-purpose multi-output GP model \cite{SLFM2005, Alvarez-2011, alvarez2012kernels} 
that uses a small number 
of $P$ latent GPs (factors) to generate the $K$ outputs through a linear mapping. The full hierarchical model for generating the training examples is,        
\begin{align}
h_p & \sim \mathcal{GP}(0, k(\bx^{(i)}, \bx^{(j)})), \ p=1,\ldots,P, \\
\bff^{(i)} & = \Phi \times \bh^{(i)} + \bb, \ i=1,\ldots,N, \\ 
\label{eq:bff}
\by^{(i)} & \sim p(\by^{(i)}|\bh^{(i)}) = \prod_{k=1}^K \sigma(y^{(i)}_k f_k^{(i)} ), 
\ i=1,\ldots,N,
\end{align}
where $\bh^{(i)} = [h_1^{(i)} \ldots, h_P^{(i)}]^\top \in \mathbb{R}^P$ denotes the vector of all 
function values evaluated at input $\bx^{(i)}$, i.e.\ $h_p^{(i)} \equiv h_p( \bx^{(i)})$,
while the parameters $\Phi \in \mathbb{R}^{K \times P}$ and $\bb \in \mathbb{R}^K$ correspond to the 
factor loadings matrix and the bias vector of the linear mapping. By using these parameters the latent vector $\bh^{(i)}$ 
is deterministically mapped into $\bff^{(i)} = [f_1^{(i)},\ldots,f_K^{(i)}]^\top \in \mathbb{R}^K$, such that each
$
f_k^{(i)} = \sum_{p=1}^P \phi_{k p} h_p^{(i)} + b_k
$
defines the so-called {\em utility} score that finally generates the $k$-th binary label through a sigmoidal/Bernoulli   
likelihood. Notice that while the labels are conditionally independent given $\bh^{(i)}$, they become 
fully coupled once these variables are integrated out. 
The full joint distribution is given by 
\begin{equation}
\prod_{i=1}^N p(\by^{(i)}|\bh^{(i)}) \prod_{p=1}^P p(\bh_p),
\label{eq:initialJoint}
\end{equation}
where $p(\bh_p) = \mathcal{N}(\bh_p | \mathbf{0}, K_X)$ is an $N$-dimensional Gaussian distribution induced by evaluating the GP prior at the training inputs $X$ with corresponding covariance matrix $K_X$. An equivalent way to write the above model is by using the concept of kernels for multi-task or vector-valued functions \cite{bonilla2008multi, Alvarez-2011, alvarez2012kernels}. More precisely, observe that the utility scores $f_k^{(i)}$ that directly interact with the data in \eqref{eq:bff} follow a GP prior with mean given by the bias $b_k$ (that depends on the label but not on the input) and covariance function
\begin{equation}
\text{Cov}(f_l^{(i)}, f_k^{(j)}) = k(\bx^{(i)}, \bx^{(j)}) \sum_{p=1}^P \phi_{l p} \phi_{k p}.
\label{eq:multitaskKernel}
\end{equation} 
For regression problems with Gaussian likelihoods the above multi-task GP is known as the intrinsic correlation model \cite{stoyan1996hans,bonilla2008multi}, a specific case of co-kriging in geostatistics; see \cite{alvarez2012kernels} for a full review. Here, we use this model for multi-label learning where the tasks correspond to different class labels. 
  
Inference in the above model is very challenging since real applications in multi-label classification
involve, very large number of training instances $N$, very large number of class labels 
$K$ and often extremely large input dimensionality $D$ \cite{Zhang2013, Gibaja2014, Gibaja2015}. 
To deal with such challenges we apply subspace inducing inputs together with  
the $\Op(M)$ parametrization of the $q(\bu)$ distribution presented in Section \ref{sec:theory}. 
We also make use of  
stochastic optimization by sub-sampling minibatches of training instances and 
possibly also of class labels to deal with very large $K$. Full details of the variational training 
procedure are given in Section \ref{sec:scalableStoch}.


\begin{table}
\caption{Data sets statistics: $N$ and $N_{ts}$ are the number of the training and test points respectively, $D$ and $K$ are the number of features and labels respectively, and $\overline{K}$ is the average number of positive labels in an instance.} 
\label{tbl_1}
\vskip 0.15in
\begin{center}
\begin{small}
\begin{tabular}{lccccr}
\toprule
Data set         & $D$       & $K$      & $N$      & $N_{ts}$   & $\overline{K}$  \\
\midrule
Bibtex           & 1836      & 159      & 4880     & 2515       & 2.40   \\
Delicious        & 500       & 983      & 12920    & 3185       & 19.03  \\
Mediamill        & 120       & 101      & 30993    & 12914      & 4.38   \\
EUR-Lex          & 5000      & 3993     & 15539    & 3809       & 5.31   \\
RCV1             & 47236     & 2456     & 623847   & 155962     & 4.79   \\
AmazonCat        & 203882    & 13330    & 1186239  & 306782     & 5.04   \\

\bottomrule
\end{tabular}
\end{small}
\end{center}
\vskip -0.1in
\end{table}

\begin{table}
\caption{Predictive Performance of the MLGPF model for the seven multi-label datasets. It is also given the computational time of both the lower bound in eq. \eqref{eq:lowerbound} and the sum of the $P$ KL divergence terms in the same bound. It can be found in the first two rows of each dataset respectively (and their corresponding derivatives). Those methods that have not reported results for a dataset are indicated with the '-' sign. 
}
\label{tbl_2}
\vskip 0.15in
\centering 
\begin{small}
\begin{tabular}{cccccccccc}
\toprule
Dataset   &       & s-{\sc linear}     &                 & {\sc linear}    &         & s-{\sc se} &   
& {\sc se}    &  \\
\midrule
          & P@1   & 59.31              & 6.7s           & \textbf{60.20}  & 6.2s   & 41.89      & 6.2s  & 38.68  &  5.7s  \\
Bibtex    & P@3   & 36.73              & 3.8s           & \textbf{37.02}  & 4.1s   & 24.30      & 4.4s  & 21.71  & 4.6s  \\
          & P@5   & \textbf{27.40}     &                 & 27.34           &         & 18.57      &        & 16.55         &  \\
\midrule
          & P@1   & 66.13              & 6.2s           & \textbf{67.08}  & 7.3s   & 59.89      & 5.8s  & 61.94  &  6.3s  \\
Delicious & P@3   & 60.38              & 4.0s           & \textbf{61.50}  & 3.9s   & 53.80      & 4.8s  & 55.91  & 4.5s   \\
          & P@5   & 55.69              &                 & \textbf{56.88}  &         & 49.21      &           & 50.88         &         \\
\midrule
          & P@1   & 82.98              & 5.4s           & 82.33           & 5.6s   & \textbf{84.12}    & 5.8s       & 82.80 & 5.0s \\
Mediamill & P@3   & 65.62              & 4.1s           & 65.25           & 3.8s   & \textbf{67.17}    & 5.0s & 66.14 & 4.7s\\
          & P@5   & 51.32              &                 & 51.09           &         & \textbf{53.15}    &        & 52.16 &      \\
\midrule
          & P@1   & \textbf{79.31}     & 9.4s           & 78.34           & 9.5s   & 66.42      & 8.6s    & 64.95 & 8.3s \\
EUR-Lex   & P@3   & \textbf{64.24}     & 4.5s           & 63.35           & 4.5s   & 50.58      & 4.5s & 49.47 & 5.1s\\
          & P@5   & \textbf{52.79}     &                 & 52.06           &         & 40.56      &        & 39.63       &      \\
\midrule
          & P@1   & \textbf{88.74}     &  8.8s          & -               & 13.3s  & 25.97      &   7.7s          & - & 11.8s        \\
RCV1      & P@3   & \textbf{71.27}     & 4.0s           & -               & 6.1s   & 21.85      &  4.3s        & - & 6.5s         \\
          & P@5   & \textbf{51.16}     &                 & -               &         & 17.13      &            & -              &         \\
\midrule
          & P@1   & \textbf{85.90}     & 22.3s          & -               &  41.6s & 44.11      &  21.5s         & - & 38.5s        \\
AmazonCat & P@3   & \textbf{64.98}     & 3.9s           & -               & 12.8s  & 27.18      & 4.8s          & - & 13.8s  \\
          & P@5   & \textbf{49.88}     &                 & -               &         & 21.18      &            & -      &        \\ 
\bottomrule
\end{tabular}
\end{small} 
\vskip -0.1in
\end{table}

\section{Experiments \label{sec:results}}

Experiments were carried out on 5 small-scale real-world datasets,  Bibtex \cite{katakis2008multilabel}, Delicious \cite{tsoumakas2008effective}, EUR-Lex \cite{mencia2008efficient}, and Mediamill \cite{snoek2006challenge}, and two  large-scale datasets,  RCV1 \cite{lewis2004rcv1} and AmazonCat \cite{mcauley2013hidden}. 
All the datasets are publicly available; see Table \ref{tbl_1} for summary statistics.

In our experiments we applied the proposed multi-label GP factor model (MLGPF) using 
a linear and a squared exponential kernel.
We either freely optimized the matrix $Z$ resulting to methods {\sc linear} and {\sc se} for linear and squared exponential kernels respectively, or we optimized the subspace inducing inputs matrix $A$ resulting to methods s-{\sc linear} and s-{\sc se}; see Tables \ref{tbl_2} and \ref{tbl_3}. Initialization of  $Z$ was achieved by running a few iterations of the k-means algorithm.  Additionally, we consider the case where inducing inputs or subspace inducing inputs are kept fixed; see Section \ref{sec:extra_results} for the corresponding results. 

We evaluated the predictive performance of our method against the golden standards by using of the  \textit{Precision@k} score (P@k). For a ground truth test vector $\by^{(*)} \in \{-1, 1\}^K$ and a predicted score vector $\bar{\bff}^{(*)} \in \mathbb{R}^K$, the P@k is defined as $k^{-1} \sum_{l \in \text{rank}_k(\bar{\bff}^{(*)})} (\by_{l}^{(*)} + 1)/2$, where $\text{rank}_k(\bar{\bff}^{(*)})$ returns the $k$ largest indices of $\bar{\bff}^{(*)}$ in descending order. Here, $\bar{\bff}^{(*)}$ can be evaluated using the trained MLGPF model as described in Section \ref{sec:prediction}.
Such ranking-based evaluation of multi-label models is very standard in the multi-label literature; see, for example,  \cite{Prabhu-2014,jain17a} and the reported results in the \textit{Extreme Classification Repository}.\footnote{ http://manikvarma.org/downloads/XC/XMLRepository.html}

In all datasets, apart from EUR-Lex, we used $P=30$ latent GP functions and $M=500$ inducing inputs.  For EUR-Lex we set, after some experimentation,  $P=40$  and $M=500$ since greater values do not improve performance.
The minibatch sizes  were set  to $|X_b| = 500$ for all datasets except EUR-Lex where we used  $|X_b| = 800$.   Bibtex was run for 400 epochs, Delicious, Mediamill, and EUR-Lex were run for 200 epochs, RCV1 was run for 20 epochs and AmazonCat for 15. 
We chose $R$ to be close to half of the dimensionality of the input space for the small-scale datasets, i.e $R=1000$ for Bibtex, $R=250$ for Delicious, $R=70$ for Mediamill, and $R=2500$ for Eurlex, while for large scale datasets we set  $R=2000$ for RCV1 and $R=2000$ for AmazonCat.

\begin{table}
\caption{Performance comparison between the MLGPF model using s-{\sc linear}  and other state-of-the-art methods. 
Those methods that have not reported results for a dataset are indicated with the '-' sign.}
\label{tbl_3}
\vskip 0.15in
\centering 
\begin{small}
\begin{tabular}{ccccccc}
\toprule 
Dataset   &       & s-{\sc linear}  & SLEEC            & PfastreXML      & FastXML          & PD-Sparse\\
\midrule
          & P@1   & 59.31          & \textbf{65.08}   & 63.46           & 63.42            & 61.29   \\
Bibtex    & P@3   & 36.73          & \textbf{39.64}   & 39.22           & 39.23            & 35.82   \\
          & P@5   & 27.40          & 28.87            & \textbf{29.14}  & 28.86            & 25.74   \\ 
\midrule
          & P@1   & 66.13          & 67.59            & 67.13           & \textbf{69.61}   & 51.82   \\
Delicious & P@3   & 60.38          & 61.38            & 62.33           & \textbf{64.12}   & 44.18   \\
          & P@5   & 55.69          & 56.56            & 58.62           & \textbf{59.27}   & 38.95   \\
\midrule
          & P@1   & 82.52          & \textbf{87.82}   & 83.98           & 84.22            & 81.86   \\
Mediamill & P@3   & 65.63          & \textbf{73.45}   & 67.37           & 67.33            & 62.52   \\
          & P@5   & 51.32          & \textbf{59.17}   & 53.02           & 53.04            & 45.11   \\
\midrule
          & P@1   & \textbf{79.31} & 79.26            & 75.45           & 71.36            & 76.43   \\
EUR-Lex   & P@3   & 64.24          & \textbf{64.30}   & 62.70           & 59.90            & 60.37   \\
          & P@5   & \textbf{52.79} & 52.33            & 52.51           & 50.39            & 49.72   \\
\midrule
          & P@1   & 88.74          & -                & -               & \textbf{91.23}   & -       \\
RCV1      & P@3   & 71.27          & -                & -               & \textbf{73.51}   & -       \\
          & P@5   & 51.16          & -                & -               & \textbf{53.31}   & -       \\
\midrule
          & P@1   & 85.90          & 90.53            & 91.75           & \textbf{93.11}   & 90.60   \\
AmazonCat & P@3   & 64.98          & 76.33            & 77.97           & \textbf{78.20}   & 75.14   \\
          & P@5   & 49.88          & 61.52            & \textbf{63.68}  & 63.41            & 60.69   \\ 
\bottomrule
\end{tabular}
\end{small} 
\vskip -0.1in
\end{table} 
For the large-scale datasets, we chose to only optimize subspace inducing inputs in order to make feasible the optimization over the extremely high dimensional input spaces. 

All the results can be found in Tables \ref{tbl_2} and \ref{tbl_3}. Table \ref{tbl_2} contains computational times of both the lower bound in eq. \eqref{eq:lowerbound} and the $P$ KL divergence terms of the same bound where each time includes the computational time of their derivatives too. All experiments were run on an Intel Xeon Processor E5-2667 v3 server.

The running times of Table \ref{tbl_2} show the considerable speed gain that we achieve using subspace inducing inputs as the input dimensionality increases. For example, computation of the KL divergence terms for the AmazonCat using  s-{\sc linear} is more that four times faster than {\sc linear}.

Regarding predictive performance of the multi-label GP factor model, we notice that we achieve close performance with the golden standards and we achieve better predictive performance in EUR-Lex dataset. Notice that in our current experiments 
we are mostly interested in showing the scalability of our sparse GP algorithm rather than improving 
the state-of-the-art in multi-label classification. Combined 
kernels and more suitable likelihood functions could have been needed to overcome state-of-the-art algorithms; we leave this for future work, see also the discussion in Section 5.   
      
Additionally, subspace inducing inputs gave better results in some occasions, revealing the optimization efficiency of utilizing subspace of variational parameters in (sparse) high dimensional spaces. These outcomes can be also justified by the corresponding evolution of the lower bounds as depicted in fig. \ref{fig:lowerbounds_opt} of Appendix. 

Further, since the performance superiority of the linear over the SE kernel is observed in most of the datasets, we compare that kernel using subspace inducing inputs with four state-of-the-art-methods from the literature, such as SLEEC  \cite{bhatia2015sparse}, PFastreXML \cite{jain2016extreme}, FastXML \cite{prabhu2014fastxml}, and  PD-Sparse \cite{yen2016pd} as they are reported in the \textit{Extreme Classification Repository} (see footnote 1). Our proposed MLGPF method using subspace inducing inputs remains very close and in some cases, such as the EUR-Lex dataset, outperforms all the baselines.

\section{Discussion \label{sec:discussion}}
 
We have presented a fully scalable sparse GP variational inference implementation framework that is useful to GP applications with large input data dimensionality.  Such datasets appear more often in modern machine learning applications and there has been a need to enrich the GP computational quiver to accommodate new, challenging problems. 

We tested our proposed framework to the challenging extreme multi-label classification problem.  We constructed a new GP factor model that induces correlations to the labels and we presented the computing efficiency and predictive performance of the GP factor model.  The results, especially because they have been compared against golden standards, seemed very satisfactory given that the GP model was not expected to perform optimally in such very sparse datasets in which non-linear GP classifiers might not be so useful.  We currently work on elaborating further  the modelling aspects of our method such as to modify the likelihood in  order to deal with missing labels \cite{jain17a} and add extra latent variables that can capture non-input dependent correlation between the class labels \cite{Gibaja2014}.  We believe that our implementation strategy will allow more flexibility for GP-based models to other interesting machine learning research areas.

\bibliographystyle{abbrv}
\bibliography{refs}

\appendix
\section{Appendix  \label{appendix}} 

\subsection{Scalable Variational Inference \label{sec:scalable}}

The approximate inference procedures derived in this section are mainly based 
on the representation that uses the latent GP vectors $\bu_p$ rather than the multi-task kernel representation in eq. (16) of the main paper. The utility scores $f_k^{(i)}$ will only be used to simplify the computations of some final Gaussian integrals. 

To deal with large number of training data we consider the variational sparse GP inference
framework based on inducing variables as described in Section 2.1 of the main paper. For each latent function $h_p$ we introduce a vector of inducing variables $\bu_p \in \mathbb{R}^M$ of function values of $h_p$ evaluated at inputs $Z = A \widetilde{X} $, where for simplicity we take the variational parameters matrix $A$ to be shared by all latent GPs. By following the same steps from Section 2.1, we augment the joint distribution 
in eq. (1) of the main paper  with the inducing variables to obtain
\begin{equation}
\prod_{i=1}^n p(\by^{(i)}|\bh^{(i)}) \prod_{p=1}^P p(\bh_p | \bu_p) p(\bu_p).
\label{eq:augmentedJoint}
\end{equation}
Here, $p(\bu_p) = \mathcal{N} (\bu_p | \mathbf{0}, K_Z)$
is the marginal GP prior over $\bu_p$ while $p(\bh_p | \bu_p)$ is the conditional GP prior which can be written as $p(\bh_p | \bu_p) = \mathcal{N} \left( \bh_p | K_{X Z} K_{Z}^{-1} \bu_p, K_X -  K_{X Z} K_{Z}^{-1} K_{Z X} \right)$. 
The approximate distribution in our case now will be the following,
\begin{equation}
\prod_{p=1}^P p(\bh_p | \bu_p) q(\bu_p), 
\label{eq:approx_dist}
\end{equation}
where $p(\bh_p | \bu_p)$ is the conditional GP prior,
while $q(\bu_p) = \mathcal{N}(\bu_p | \bm^u_p, \bS^u_p)$ is a Gaussian variational distribution 
over the inducing variables for the $p$-th latent GP with $(\bm^u_p, \bS^u_p)$ parametrized as follows, 
\begin{align}
\bm^u_p & = K_Z \bmu_p, \nonumber \\ 
\bS^u_p &  = (K_Z^{-1} + \Sigma_p^{-1})^{-1} = K_Z - K_Z (K_Z + \Sigma_p)^{-1} K_Z. \nonumber
\end{align}
This parametrization of $q(\bu_p)$ is one of the novelties of our method. Specifically,  
$\bmu_p \in \mathbb{R}^{M}$ is a real-valued vector of tunable variational parameters 
and $\Sigma_p$ is a diagonal positive definite matrix (i.e.\ with each diagonal element restricted to be non-negative) parametrized by $M$ variational parameters needed for the diagonal elements. Overall 
$q(\bu_p)$ is parametrized by $2 M$ 
variational parameters 
while all the remaining structure comes from a careful preconditioning with the model kernel matrix $K_Z$ that appears in  
the GP prior over $\bu_p$. 
The parametrization of the mean $\bm^u_p = K_Z \bmu_p$ has been used before for full GPs \cite{opper2009variational,Damianou11}, and it allows to speed up optimization, since $\bmu_p$ tends be more noisy (and therefore much more easily optimizable) than the smoother $\bm^u_p$. 
The specific choice of the covariance matrix, $\bS^u_p  = (K_Z^{-1} + \Sigma_p^{-1})^{-1}$, 
mimics the structure of the covariance matrix of the optimal $q^*(\bh_p)$ obtained by a non-sparse (i.e.\ without using inducing variables) approximation associated with imposing the factorized approximation $\prod_{p=1}^P q^*(\bh_p)$. Specifically, an optimal $q^*(\bh_p)$ has covariance $(K_X^{-1} + \Lambda_p)^{-1}$ where $\Lambda_p$ is a diagonal 
positive definite matrix; see Appendix \ref{sec:opt_covariance} for a 
proof that follows the derivations in \cite{opper2009variational}. The corresponding $q(\bh_p)$ obtained by the previous choice of $q(\bu_p)$ (derived by marginalizing out 
$\bu_p$ from the variational distribution in \eqref{eq:approx_dist}) is 
$q(\bh_p) = \mathcal{N}(\bh_p | \bm^h_p, \bS^h_p)$ where
\begin{align}
\bm^h_p & = K_X \bmu_p, 
\label{eq:mph}
 \\ 
\bS^h_p & = K_X - K_{X Z} (K_Z + \Sigma_p)^{-1} K_{Z X}.
\label{eq:Sph}
\end{align}
This parametrization can recover the optimal $q^*(\bh_p)$ when we place the inducing inputs on the
training inputs, i.e.\ when $Z=X$ or in our case when $A = U S$. In other cases the above covariance matrix of $q(\bh_p)$ will not be able to match 
exactly the optimal one of $q^*(\bh_p)$, but still in practice it tends to be very flexible especially when we optimize over
the inducing inputs $Z$ so that a posteriori $\bh_p$ is well reconstructed by $\bu_p$ (i.e.\ 
when $p(\bh_p | \bu_p)$ has low entropy). 

Two important benefits associated with the above parametrization of $q(\bu_p)$ are: (i) 
it reduces the number of extra variational parameters to $\Op(M)$ (rather than $\Op(M^2)$)  
while still remaining very flexible and (ii) through the preconditioning with the matrix $K_{Z}$ 
it leads to a numerically stable and simplified form of the lower bound as shown next. 

To express the lower bound on the log marginal likelihood $\log p(Y)$   
under the variational distribution in \eqref{eq:approx_dist} we start the derivation as in Section 2.1 of the main paper which leads to cancellation of each conditional GP prior $p(\bh_p | \bu_p)$. Then, by following the derivation  suitable for scalable and/or non-Gaussian likelihoods \cite{hensman2013gaussian,Lloyd2015,hensman2015scalable, Dezfouli2015} and using the lower bound of eq. (3) of the main paper, we obtain (see Section for the derivation of the lower bound ),
\begin{align}
\mathcal{F} & = - \sum_{i=1}^N  \sum_{k=1}^K \mathbb{E}_{q(f_k^{(i)})}  \left[ \log ( 1 + e^{-y_{k}^{(i)} f_k^{(i)} })  \right] \nonumber  \\
& -  \sum_{p=1}^P \text{KL}[ q(\bu_p) || p(\bu_p) ].
\label{eq:lowerbound} 
\end{align}
In the first line of this expression we have written the expectation of each log-likelihood term as an integral 
under the scalar utility $f_k^{(i)} = \sum_{p=1}^P \phi_{kp} h_p^{(i)} + b_k$, that follows the univariate variational Gaussian distribution  
\begin{equation}
q(  f_k^{(i)} )   = \mathcal{N}( f_k^{(i)}   | \sum_{p=1}^P \phi_{kp} m_p^{(i)} + b_k, 
\sum_{p=1}^P \phi_{kp}^2 s_p^{(i)} ),
\label{eq:qfk}
\end{equation}
where $m_p^{(i)}$ is the $i$-th element of the vector $\bm^h_p$ defined in \eqref{eq:mph} and $s_p^{(i)}$ 
the $i$-th diagonal element (i.e.\ variance) of the covariance matrix $\bS^h_p$ from \eqref{eq:Sph}. Clearly, all 
expectations over the likelihood terms reduce to performing
$N K$ one-dimensional integrals under Gaussian distributions and each such integral can be  accurately approximated 
by Gaussian quadrature. 

Each KL divergence term of the lower bound in the second line of eq.\ \eqref{eq:lowerbound} is given by
\begin{align}
&  \text{KL} [q(\bu_p) || p(\bu_p)] = \frac{1}{2} \bmu_p^\top K_Z \bmu_p - 
\frac{1}{2} \tr \left(  (K_Z + \Sigma_p)^{-1} K_Z \right) \nonumber \\
& + \frac{1}{2} \log |K_Z + \Sigma_p| - \frac{1}{2} \log |\Sigma_p|. 
\end{align}
Notice that this term and the overall bound in \eqref{eq:lowerbound} has a quite simplified 
and numerically stable form. This is because the chosen parametrization of $q(\bu_p)$ leads to cancellation of all 
inverses and determinants of $K_Z$. Thus, unlikely other sparse GP lower bounds including the optimal one 
in GP regression \cite{titsias2009variational}, the above bound does not require  
the computation of the Cholesky decomposition of $K_Z$, which requires 
"jitter" addition to be numerically stable. Instead, the  matrix we need to decompose using Cholesky is $K_Z + \Sigma_p$, which is in an already numerically stable form due to the inflation of the diagonal of $K_Z$ with $\Sigma_p$. In a practical implementation 
 we can constrain the diagonal variational parameters of $\Sigma_p$ 
to be larger than a small value (typically $10^{-6}$) to ensure numerical stability throughout optimization.

To compute the bound we need firstly to perform $P$ Cholesky decompositions of the matrices $K_Z + \Sigma_p$ 
that overall scales as $\Op(P M^3)$ and allows us to fully compute the sum of the KL divergence terms in the second line in 
\eqref{eq:lowerbound}. Notice that the use of the parametrization $Z$ using $A$ allows us to compute $K_Z$ in $\Op(M^3)$ otherwise even the $\Op(P M^3)$ term would be dominated in practice by the $\Op(D M^2)$ for extremely large $D$. Then, with these Cholesky decompositions 
precomputed, for each $i$-th data point we need to compute $(m_p^{(i)},s_p^{(i)})_{p=1}^P$, an operation 
that scales as $\Op(P M^2)$, and subsequently compute the $K$ variational distributions  (i.e.\ their means and variances) over the utility scores in \eqref{eq:qfk} which requires additional $\Op(K P)$ time. Therefore, in order to 
compute the whole data reconstruction term of the bound
(first line in eq.\ \eqref{eq:lowerbound}) we need $\Op(N K P + N P M^2)$ time
and for the full bound we need  $\Op(N K P + N P M^2 + P M^3)$ time.
Given that $N \gg M$ and $K \gg P$, 
the terms that can dominate are either $\Op(N K P)$ or  $\Op(N P M^2)$ which can make the computations very expensive 
when the number of data instances and/or labels is very large. Next, we show how to make the optimization of
 the bound scalable for arbitrarily large numbers of data points and labels.

\subsection{Scalable Training using Stochastic Optimization \label{sec:scalableStoch}}

To ensure that the time complexity $\Op(N K P + N P M^2 + P M^3)$ for very large datasets is reduced to  
$\Op(P M^3)$ we shall optimize the bound using stochastic gradient 
ascent by following a similar procedure used in stochastic variational inference 
for GPs \cite{hensman2013gaussian}. Given that the sum of 
KL divergences in \eqref{eq:lowerbound} is already within the desired complexity
$\Op(P M^3)$, we only need to speed up 
the remaining data reconstruction term. This term  
involves a double sum over data instances and class labels, a setting suitable for stochastic approximation.
Thus, a straightforward procedure is to uniformly sub-sample terms 
in the double sum in \eqref{eq:lowerbound} which leads to an unbiased estimate of the bound 
and its gradients. In turns out that we can further reduce the variance 
of this basic strategy by applying a more stratified sub-sampling 
over class labels as discussed next. 

Suppose $\bB \subset \{1,\ldots,N\}$ denotes the current minibatch at the $t$-th iteration of stochastic gradient ascent. 
For each $i \in \bB$ the internal sum over class labels can be written as  
$$
{
-\sum\limits_{k \in \mathcal{P}_{i}}\mathbb{E}_{q(f_k^{(i)})} \log( 1 + e^{- f_k^{(i)} })
-\sum\limits_{\ell \in \mathcal{N}_{i}} \mathbb{E}_{q(f_{\ell}^{(i)})} \log( 1 + e^{f_{\ell}^{(i)} }) 
}
$$
where $\mathcal{P}_i = \{ k | y_{k}^{(i)} = 1 \}$ is the set of present or
positive labels of $\bx^{(i)}$ while 
$\mathcal{N}_i = \{ k | y_{k}^{(i)} = - 1 \}$ is the set of absent or negative 
labels such that  $\mathcal{P}_i \cup \mathcal{N}_i = \{ 1, \cdots, K \}$. 
In typical multi-label classification problems \cite{Zhang2013, Gibaja2014, Gibaja2015} the size of positive labels  
$\mathcal{P}_i$ is very small, while the negative set can be 
extremely large. Thus, we can enumerate exactly the first sum 
and use (if needed) sub-sampling to approximate the second 
sum over the negative labels. The whole process becomes somehow similar
to negative sampling used in large scale classification and for 
learning word embeddings \cite{mikolov2013}. Overall, we get the 
following unbiased stochastic estimate of the lower bound, 
\begin{align}
& - \frac{N}{ | \mathcal{B} | } \ \sum_{i \in \mathcal{B}} \left[ 
 \sum_{k \in \mathcal{P}_i} \mathbb{E}_{q(f_k^{(i)})} \log ( 1 + e^{-f_k^{(i)} })  + \right. \nonumber \\ 
& \left. \frac{|\mathcal{N}_i|}{|\mathcal{L}_i|} \sum\limits_{\ell \in \mathcal{L}_i} \mathbb{E}_{q(f_{\ell}^{(i)})} \log ( 1 + e^{f_{\ell}^{(i)} }) \right]  
-  \sum_{p=1}^P \ \text{KL}[ q(\bu_p) || p(\bu_p) ], 
\label{eq:stochlowerbound}
\end{align}
where $\mathcal{L}_i$ is the set of negative classes for the $i$-th data 
point. 
In general, the computation of this stochastic bound scales as $\Op( |\mathcal{B} | ( |\mathcal{P}_i| + |\mathcal{L}_i|) P + |\mathcal{B} | P M^2 + P M^3)$ and by choosing $|\mathcal{B} | \sim \Op(M)$ and  $|\mathcal{P}_i| + |\mathcal{L}_i| \sim \Op(M^2)$ we can ensure that the overall time is $\Op(P M^3)$. Notice that the second condition is not that restrictive and in many cases might not be needed, i.e.\ in practice we can use very large negative sets $\mathcal{L}_i$ which for many datasets could be equal to the full negative set $\mathcal{N}_i$.    
  
We implemented the above stochastic bound in Python (where the one-dimensional integrals are obtained by 
Gaussian quadrature) in order to jointly optimize using stochastic gradient ascent and automatic differentiation tools\footnote{We used autograd: https://github.com/HIPS/autograd.}
over the parameters $(\Phi, \bb)$ of the linear mapping, the $2 P M$ variational parameters  $\{ \bmu_p, \text{diag}(\Sigma)_p \}_{p=1} ^P$ of the variational distributions $q(\bu_p)$, the inducing inputs $Z$ and the kernel hyperparameters $\btheta$.

\subsection{Prediction \label{sec:prediction}}

Given a novel data point $\bx^{(*)}$ we would like to make prediction over its unknown label vector $\by^{(*)}$.
This requires approximating the predictive distribution $p(\by^{(*)}|Y)$, 
\begin{equation}
p(\by^{(*)}|Y) \approx \int p(\by^{(*)} | \bu^{(*)})  q(\bu^{(*)}) d \bu^{(*)}.
\label{eq:predictive}
\end{equation}      
Here, $q( \bu^{(*)})$ is the variational predictive posterior over the latent function values $\bu^{(*)}$ 
evaluated at $\bx^{(*)}$. An interesting aspect of the variational sparse GP method 
is that to obtain $q( \bu^{(*)})$ we need to make no further  
approximations since everything follows from the GP consistency property, i.e.\ 
\begin{align*}
q(\bu^{(*)}) & = \prod_{p=1}^P \int p( u^{(*)}_p |\bh_p, \bu_p) 
p(\bh_p | \bu_p ) q(\bu_p) d \bh_p d \bu_p \\ 
& = \prod_{p=1}^P \int p( u^{(*)}_p | \bu_p) q(\bu_p) d \bu_p = 
\prod_{p=1}^P  q(u^{(*)}_p). 
\label{eq:qustar}
\end{align*}
Here, GP consistency tractably simplifies each integral $\int p( u^{(*)}_p |\bh_p, \bu_p) 
p(\bh_p | \bu_p ) d \bh_p = p( u^{(*)}_p | \bu_p)$ so that the obtained $p( u^{(*)}_p | \bu_p)$ is 
the conditional GP prior 
of $u^{(*)}_p$ given the inducing variables. The final form of each univariate Gaussian 
$q(u^{(*)}_p)$ has a mean and variance given precisely by equations  (17) 
and (18) from the main paper with $X$ replaced by $\bx^{(*)}$. In practice, when we compute 
several accuracy ranking-based scores that are often used in the literature to report multi-label classification 
performance \cite{Zhang2013, Gibaja2014, Gibaja2015}  it suffices to further approximate $q(\bu^{(*)})$ by a delta mass centred at the MAP\footnote{Estimating such accuracy scores using a more accurate Monte 
Carlo estimation of \eqref{eq:predictive} leads to very similar results.}. This reduces 
the whole computation of such scores to only requiring the evaluation of 
the mean utility vector $\bar{\bff}^{(*)} = [\bar{f}_1^{(*)} \ldots \bar{f}_K^{(*)}]^\top$    
such that 
$\bar{f}_k^{(*)} = \sum_{p=1}^P \phi_{k p} m_p^{(*)} + b_k$,
where $m_p^{(*)} = \bk(\bx^{(*)}, Z) \bmu_p$ and  $\bk(\bx^{(*)}, Z)$ is the cross covariance row 
vector between $\bx^{(*)}$ and the inducing points $Z$. By using $\bar{\bff}^{(*)}$ we can compute several ranking scores as described in the Results Section of the main paper.

\subsection{Derivation of the lower bound}

Here we show the steps of the derivation of the bound in eq.\ (19) in the main paper. Recall 
that for the augmented joint distribution 
\begin{equation}
\prod_{i=1}^N p(\by^{(i)}|\bh^{(i)}) \prod_{p=1}^P p(\bh_p | \bu_p) p(\bu_p), \nonumber 
\end{equation}
we would like to approximate the true posterior $ \mathcal{P} \equiv p( \{\bh_p,\bu_p\}_{p=1}^P |Y)$ with the following 
variational distribution 
\begin{equation}
\mathcal{Q} = \prod_{p=1}^P p(\bh_p | \bu_p) q(\bu_p), 
\label{eq:vardist}
\end{equation}
The minimization of the KL divergence  $\text{KL}[\mathcal{Q} ||  \mathcal{P}]$ is equivalently 
expressed as the maximization of the following lower bound on the log marginal likelihood $\log p(Y)$, 
\begin{align}
& \mathbb{E}_{ \mathcal{Q} } \left[ \log 
\frac{\prod_{i=1}^N p(\by^{(i)}|\bh^{(i)}) \prod_{p=1}^P p(\bh_p | \bu_p) p(\bu_p)}
{\prod_{p=1}^P p(\bh_p | \bu_p) q(\bu_p)} \right] \nonumber \\
& \mathbb{E}_{ \mathcal{Q} } \left[ \log 
\frac{\prod_{i=1}^N p(\by^{(i)}|\bh^{(i)}) \prod_{p=1}^P p(\bu_p)}
{\prod_{p=1}^P q(\bu_p)} \right] \nonumber \\
& \sum_{i=1}^N \mathbb{E}_{ \mathcal{Q} }\left[ \log p(\by^{(i)}|\bh^{(i)}) \right] 
- \sum_{p=1}^P \mathbb{E}_{ \mathcal{Q} }\left[ \log \frac{q(\bu_p)}{ p(\bu_p)} \right] \nonumber 
\end{align}
Since each $\mathcal{Q}$ is given by \eqref{eq:vardist}, each term in the second sum simplifies to become an expectation
over $\bu_p$, 
$$
 \mathbb{E}_{ \mathcal{Q} }\left[ \log \frac{q(\bu_p)}{ p(\bu_p)} \right]   
= \mathbb{E}_{ q(\bu_p) }\left[ \log \frac{q(\bu_p)}{ p(\bu_p)} \right] 
$$
which is precisely the KL divergence $\text{KL}[q(\bu_p) || p(\bu_p)]$. Regarding each $i$-th term in the first sum 
we first observe that 
\begin{align}
 \log p(\by^{(i)}|\bh^{(i)}) & = \sum_{k=1}^K \log \sigma( y^{(i)}_k f_k^{(i)}) \nonumber \\
 & = - \sum_{k=1}^K \log( 1  +  e^{- y^{(i)}_k f_k^{(i)} } ) \nonumber
\end{align}
where $f_k^{(i)} = \sum_{p=1}^P \phi_{k p} h_p^{(i)} + b_k$ is a scalar random variable
that under $\mathcal{Q}$ follows the univariate Gaussian distribution 
$$
q(  f_k^{(i)} )   = \mathcal{N}( f_k^{(i)}   | \sum_{p=1}^P \phi_{kp} m_p^{(i)} + b_k, 
\sum_{p=1}^P \phi_{kp}^2 s_p^{(i)} ) 
$$
where $m_p^{(i)}$ and $s_p^{(i)}$ are the mean and variance of the univariate Gaussian 
$q(h_p^{(i)}) = \int p(h_p^{(i)} | \bu_p) q(\bu_p) d  \bu_p$, 
\begin{align}
m_p^{(i)} & = \bk(\bx^{(i)}, Z) \bmu_p, \nonumber \\ 
s_p^{(i)} & = k(\bx^{(i)}, \bx^{(i)})  - \bk(\bx^{(i)}, Z) (K_Z + \Sigma_p)^{-1} \bk(Z, \bx^{(i)}). \nonumber 
\end{align}
Therefore, the whole data reconstruction term  
$\sum_{i=1}^N \mathbb{E}_{ \mathcal{Q} }\left[ \log p(\by^{(i)}|\bh^{(i)}) \right]$ 
simplifies to 
$$
 - \sum_{i=1}^N  \sum_{k=1}^K \mathbb{E}_{q(f_k^{(i)})}  \left[ \log ( 1 + e^{-y_{k}^{(i)} f_k^{(i)} })  \right]
$$

\subsection{Covariance of the optimal variational distribution $q^* (\bh_p)$ \label{sec:opt_covariance}}
\begin{proof}
We follow the proof of   
\cite{opper2009variational}. Assume that we have the factorized variational distribution 
$$\prod_{p=1}^P q(\bh_p)$$ 
where $q(\bh_p) = \mathcal{N}(\bh_p | \bm^h_p,\bS^h_p)$. 
The variational lower bound is 
\begin{align}
& =  - \sum_{i=1}^N  \sum_{k=1}^K \mathbb{E}_{q(f_k^{(i)})}  [ \log ( 1 + e^{-y_{k}^{(i)} f_k^{(i)} })  ] \nonumber \\
& - \sum_{p=1}^P \text{KL}[ q(\bh_p) || p(\bh_p) ], \nonumber
\end{align} 

where each KL divergence term is given by
\begin{align}
& \text{KL} [q( \bh_p ) || p( \bh_p )] =  \frac{1}{2} [ \tr \left( K_X^{-1} ( \bS^h_p +  \bm_p^h  \bm_p^{h \top} ) \right) \nonumber \\
& + \log |K_X| - \log |\bS^h_p|  - N ].  \nonumber 
\end{align}

Rewriting now the bound by defining the term $$V_i =  \sum_{k=1}^K \mathbb{E}_{q(f_k^{(i)})}  [ \log ( 1 + e^{-y_{k}^{(i)} f_k^{(i)} }), \quad i=1, \cdots, n ,$$ we get

\begin{equation}
\mathcal{F} (\nu) = - \sum_{i=1}^N  V_i - \sum_{p=1}^P \text{KL}[ q(\bh_p) || p(\bh_p) ]. \nonumber 
\end{equation}

Notice that each term $V_i$ is a sum of $K$ univariate Gaussian expectations with respect to the marginal $q( f_k^{(i)} )   = \mathcal{N}( f_k^{(i)}   | \sum_{p=1}^P \phi_{kp} m_p^{(i)} + b_k, 
\sum_{p=1}^P \phi_{kp}^2 s_p^{(i)} )$ which means that these expectations depend only on the linear combination of the $P$ means $m_p^{(i)}$ and the $P$ variances $s_p^{(i)}$ i.e. the  $i$-th diagonal elements of each covariance matrix $\bS^h_p$.

Therefore, by differentiating the variational lower bound with respect to each $\bS^h_p$ and setting it equal to zero we have for the covariance of the optimal variational distribution $q^*(\bh_p)$ that
\begin{align}
\label{eq:derivative_bound}
& \nabla_{\bS^h_p} \mathcal{F}(\nu) =  - \sum_{i=1}^N \nabla_{\bS^h_p} V_i  - \frac{1}{2} ( K_X^{-1} - \bS^h_p ) = 0 \nonumber \\ & \Rightarrow \bS^h_p = ( K_X^{-1} + \Lambda_p )^{-1}, \nonumber 
\end{align}

where $\Lambda_p \in \mathbb{R}^{N \times N}$ is a diagonal matrix with positive entries $\lambda_p^{(i)} = 2 \frac{\partial V_i}{\partial s_p^{(i)} } $ and for the rhs of the first line of the previous equation we made use of the matrix calculus identities, $\frac{\partial \tr ( A X )}{\partial X} = A$ and $\frac{\partial \log |X|}{\partial X} = X^{-1}$.
\end{proof}

\begin{figure}
\centering
\begin{tabular}{ccc}
{\includegraphics[scale=0.3]
{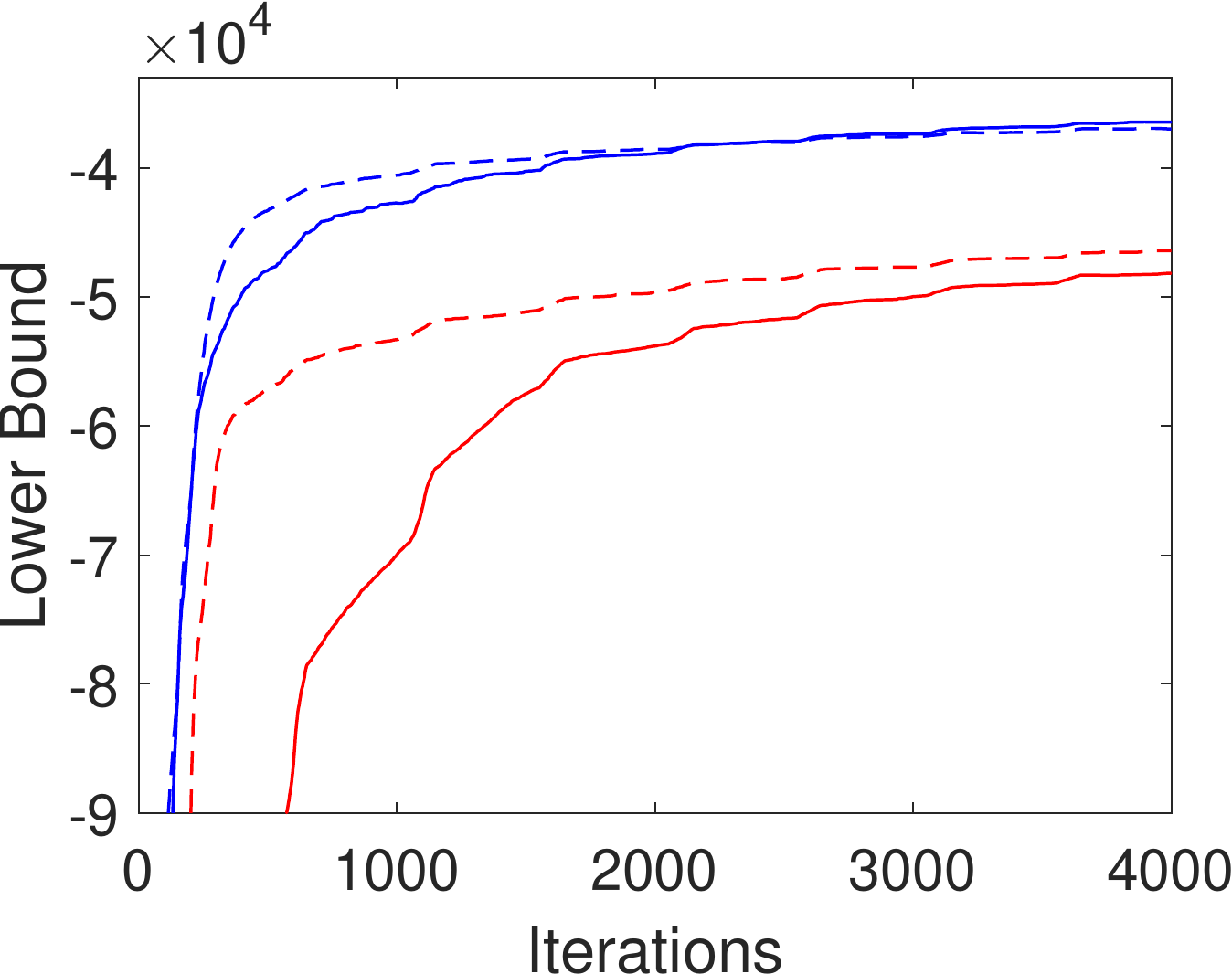}} &
{\includegraphics[scale=0.3]
{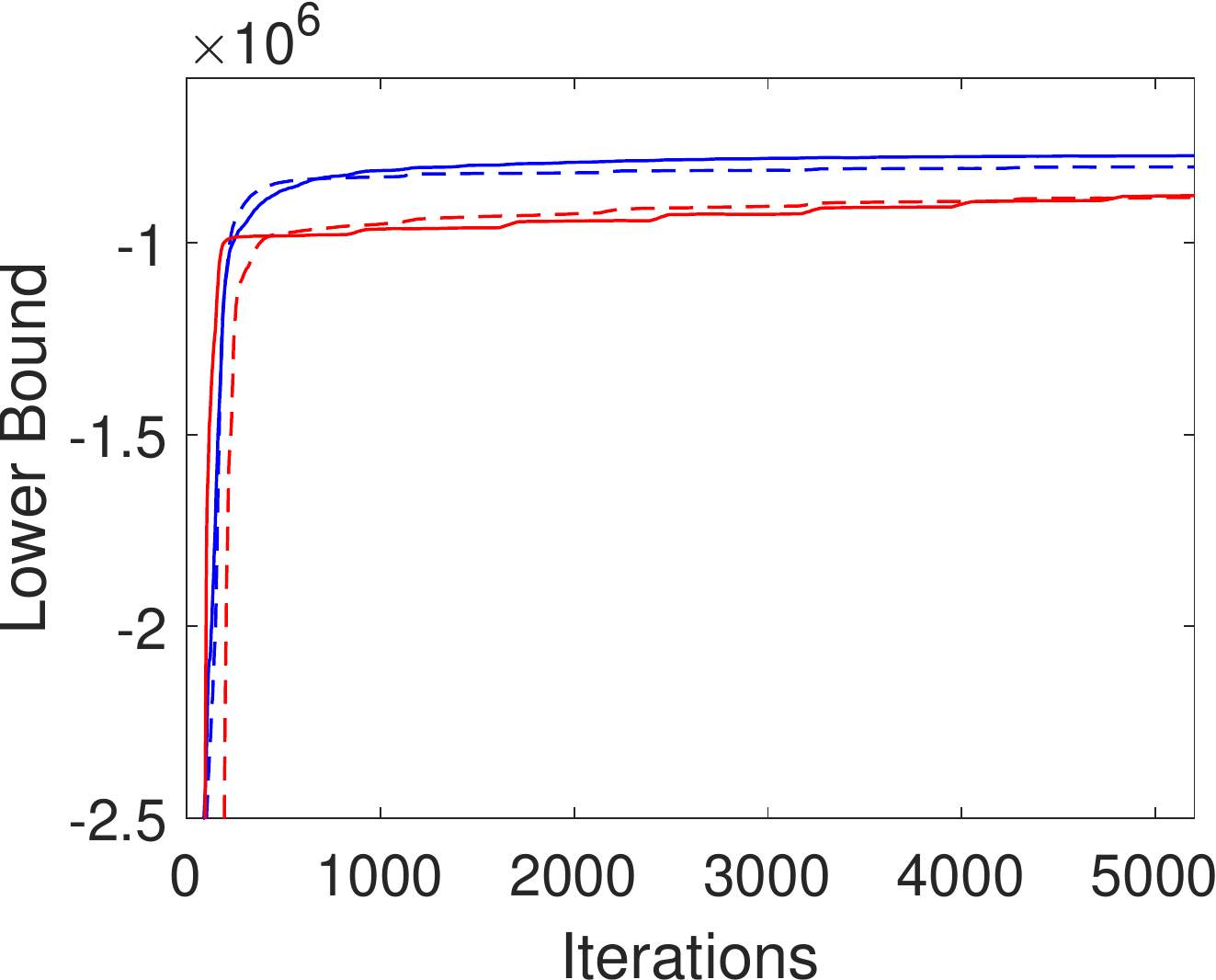}} &  
{\includegraphics[scale=0.3]  
{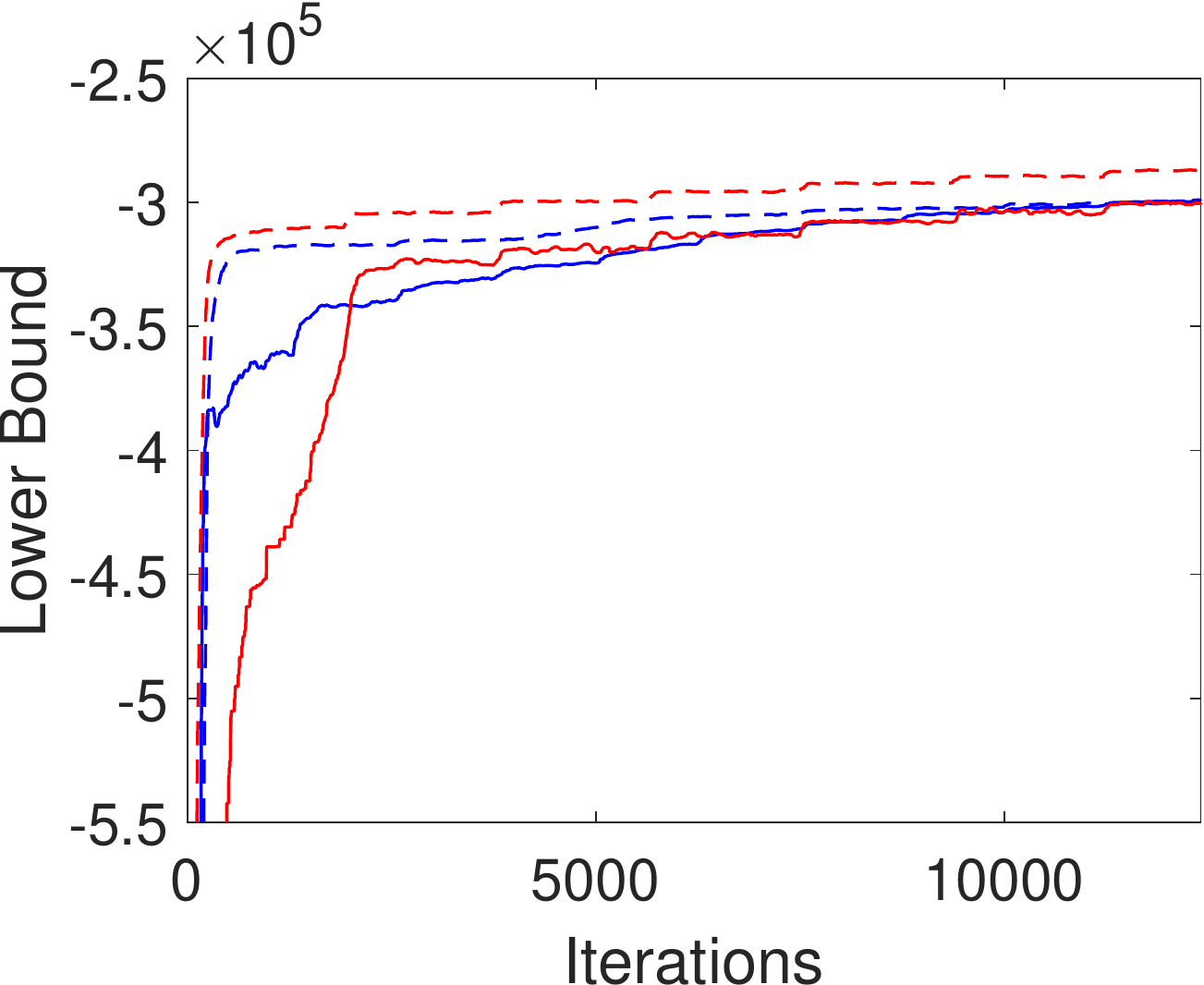}} \\
(a) & (b) & (c) \\
{\includegraphics[scale=0.3]  
{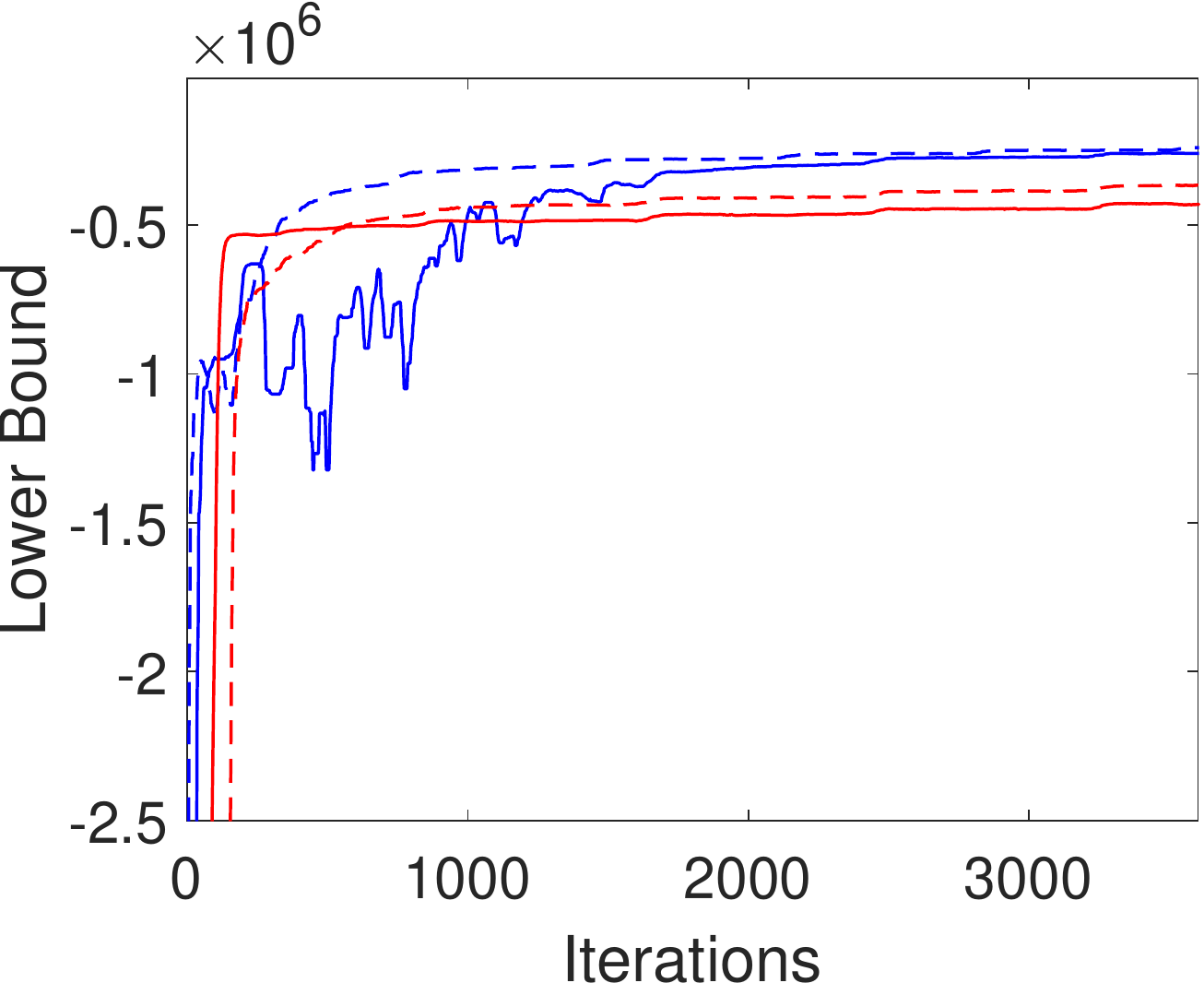}} &  
{\includegraphics[scale=0.3]  
{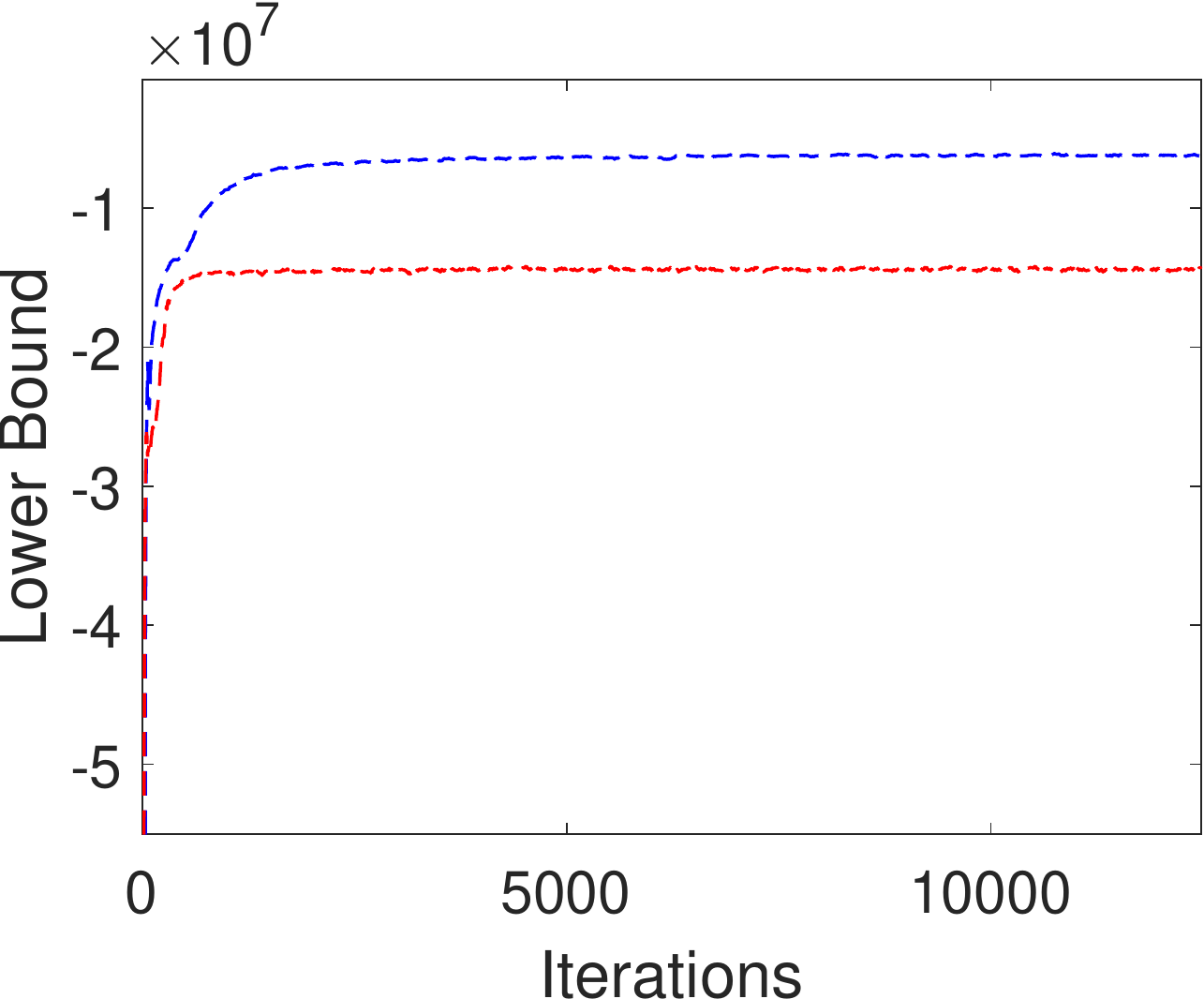}}  &
{\includegraphics[scale=0.3]  
{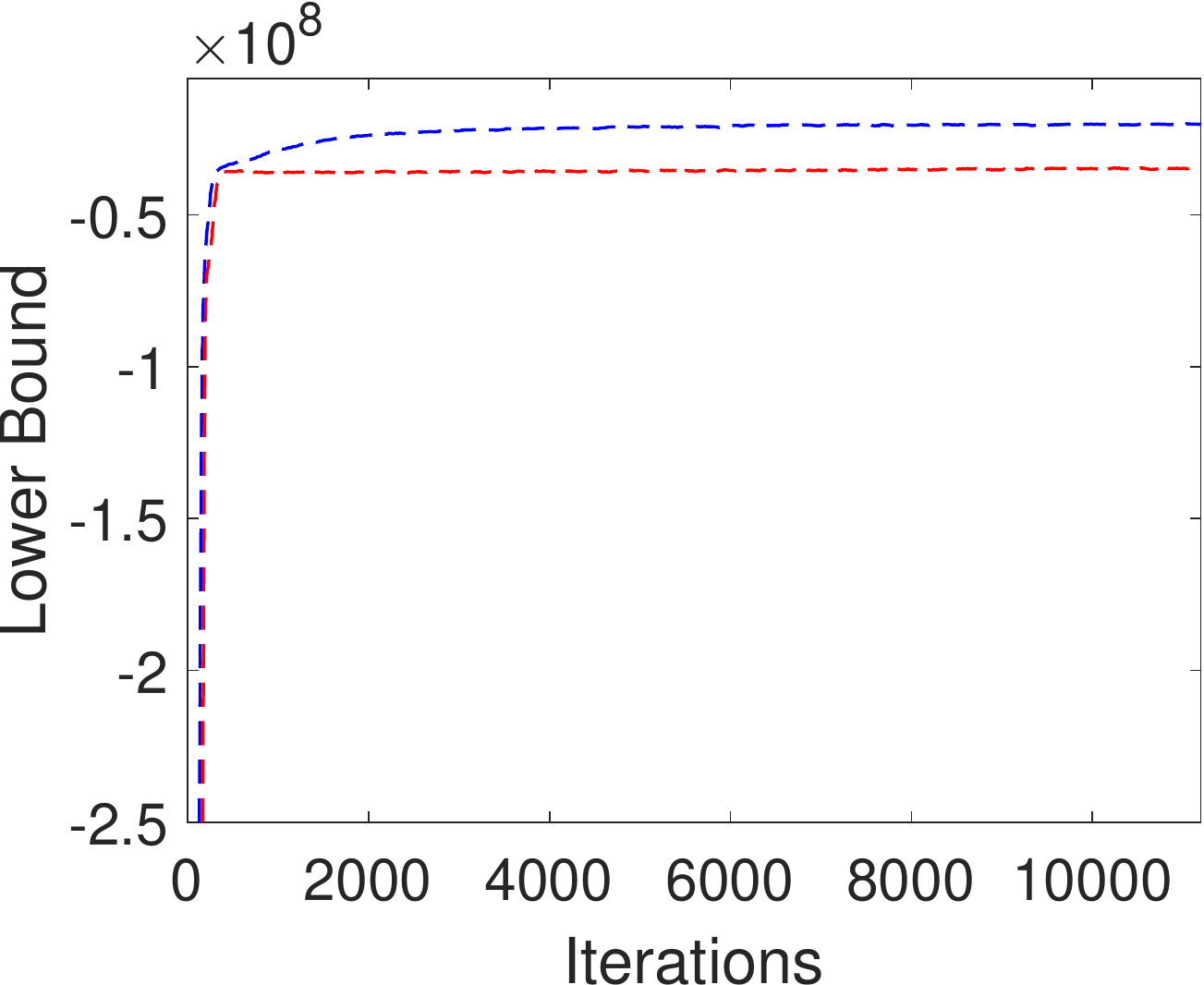}} \\
(d) & (e) & (f)
\end{tabular}
\caption{Lower bounds of (a) Bibtex, (b) Delicious, (c) Mediamill, (d) EUR-Lex, (e) RCV1, and (f) AmazonCat. The solid lines correspond to the methods that optimized the inducing points while the dashed ones correspond to those that optimized the subspace inducing points.  Blue color suggests the use of linear kernel while the red one the use of squared exponential kernel.} 
\label{fig:lowerbounds_opt}
\end{figure}

\subsection{The squared exponential case \label{sec:sqe}} 

We shall show how the representation trick of Section 2.2 can be employed in the case of squared exponential kernel, i.e. 
$$
k_{SE}(\bx, \by) = \sigma^2 \exp (  - \frac{|| \bx - \by ||^2}{2 \ell^2}), 
$$ 
where $|| \cdot ||$ is the euclidean norm. The $M \times M$ kernel matrix $K_Z$ is given by
\begin{equation}\label{eq:se_kernel}
K_Z = k_{SE}(Z, Z) =  \sigma^2 \exp (  - \frac{DZ}{2 \ell^2}),
\end{equation} 
\noindent
where $DZ$ is an $M \times M$ matrix defined as 
$$
DZ = \text{DG}_Z \mathbf{1}_M^{\top} +  \mathbf{1}_M  \text{DG}_Z^{\top} - 2 K_Z^{LIN},
$$
Here, $K_Z^{LIN}$ is given by eq.\ (7) of the main paper, $\text{DG}_Z \in \mathbb{R}^M$ includes the elements of the main diagonal of $K_Z^{LIN}$ , and  $\mathbf{1}_M$ is an $M$-dimensional column vector containing ones. The exponential term in eq. \ref{eq:se_kernel} implies element-wise exponentiation over the elements of matrix $DZ$. We can notice that the whole computational time is based on the computation of the $K_Z^{LIN}$ where we showed in Section 2.2 that it scales as $\Op(M^3)$ instead of $\Op(DM^2)$ assuming that $R \sim \Op (M)$.

Similarly, the cross covariance matrix between a minibatch of inputs $X_b$ and $Z$ can be computed as following, 
$$
K_{X_b Z} = k_{SE}(X_b, Z) =  \sigma^2 \exp (  - \frac{DZ}{2 \ell^2}),
$$
where $DZ$ now is an $|X_b| \times M$ matrix defined as 
$$
DZ = \text{DG}_{X_b} \mathbf{1}_M^{\top} +  \mathbf{1}_{|X_b|}  \text{DG}_Z^{\top} - 2 X_b Z^{\top}
$$
with  $\text{DG}_{X_b} \in \mathbb{R}^{|X_b|}$ being the main diagonal of the matrix $X_b X_b^{\top}$. Notice that the computation of the first two terms of $DZ$ scales as $\Op( M^2 + |X_b|^2 )$ while the last term given by eq.\ (8) of the main paper  scales as $\Op(|X_b| D M)$. However, as we mentioned in Section 2.2 of the main paper, this computation is fast due to the sparsity of matrix $X_b$.


\begin{table}
\caption{Predictive Performance of the MLGPF model for the seven multi-label datasets. Those methods that have not reported results for a dataset are indicated with the '-' sign.}
\label{tbl_1_appendix}
\vskip 0.15in
\centering
\begin{tabular}{cccccc}
\toprule 
Dataset   &       & sf-{\sc linear} & f-{\sc linear}  & sf-{\sc se}   & f-{\sc se}  \\
\midrule
          & P@1   & 45.12           & 40.31            & 37.25         & 36.43        \\
Bibtex    & P@3   & 26.79           & 23.16            & 20.07         & 19.42        \\
          & P@5   & 20.40           & 17.67            & 15.37         & 14.74        \\ 
\midrule
          & P@1   & 63.13           & 63.04            & 55.65         & 54.44        \\
Delicious & P@3   & 57.04           & 57.03            & 49.87         & 48.56        \\
          & P@5   & 52.26           & 52.40            & 45.62         & 44.77        \\
\midrule
          & P@1   & 75.17           & 78.75            & 82.99         & 82.69        \\
Mediamill & P@3   & 58.88           & 62.06            & 66.22         & 65.85        \\
          & P@5   & 45.33           & 47.45            & 52.26         & 51.72        \\
\midrule
          & P@1   & 70.10           & 70.70            & 57.23         & 31.32        \\
EUR-Lex   & P@3   & 53.86           & 54.07            & 42.76         & 22.49        \\
          & P@5   & 43.15           & 43.62            & 34.08         & 18.06        \\
\midrule
          & P@1   & 43.19           & -                & 30.61         & -            \\
AmazonCat & P@3   & 25.29           & -                & 19.14         & -            \\
          & P@5   & 20.66           & -                & 11.64         & -            \\ 
\bottomrule
\end{tabular}
\vskip -0.1in
\end{table}

\begin{figure}
\centering
\begin{tabular}{ccc}
{\includegraphics[scale=0.3]
{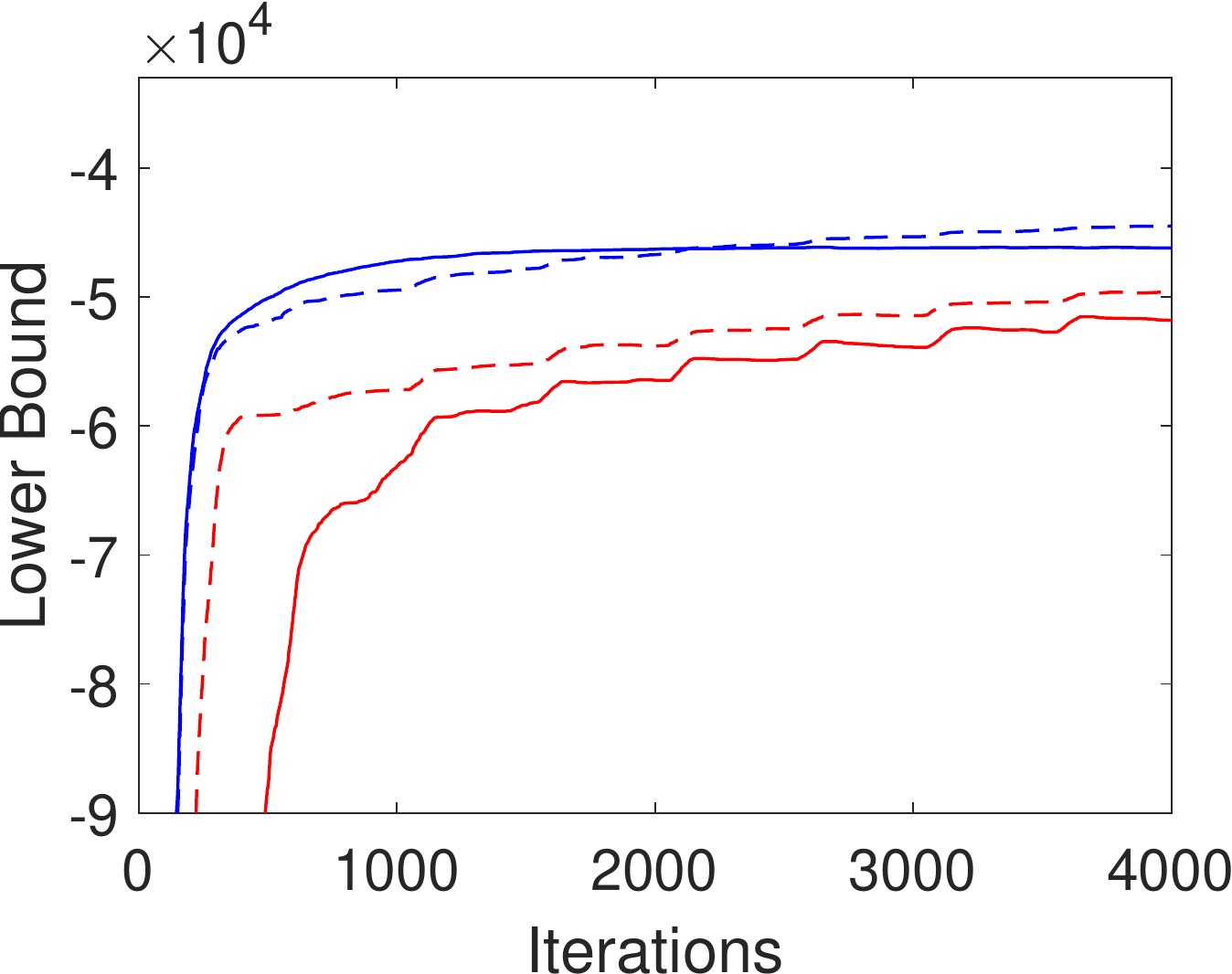}} &
{\includegraphics[scale=0.3]
{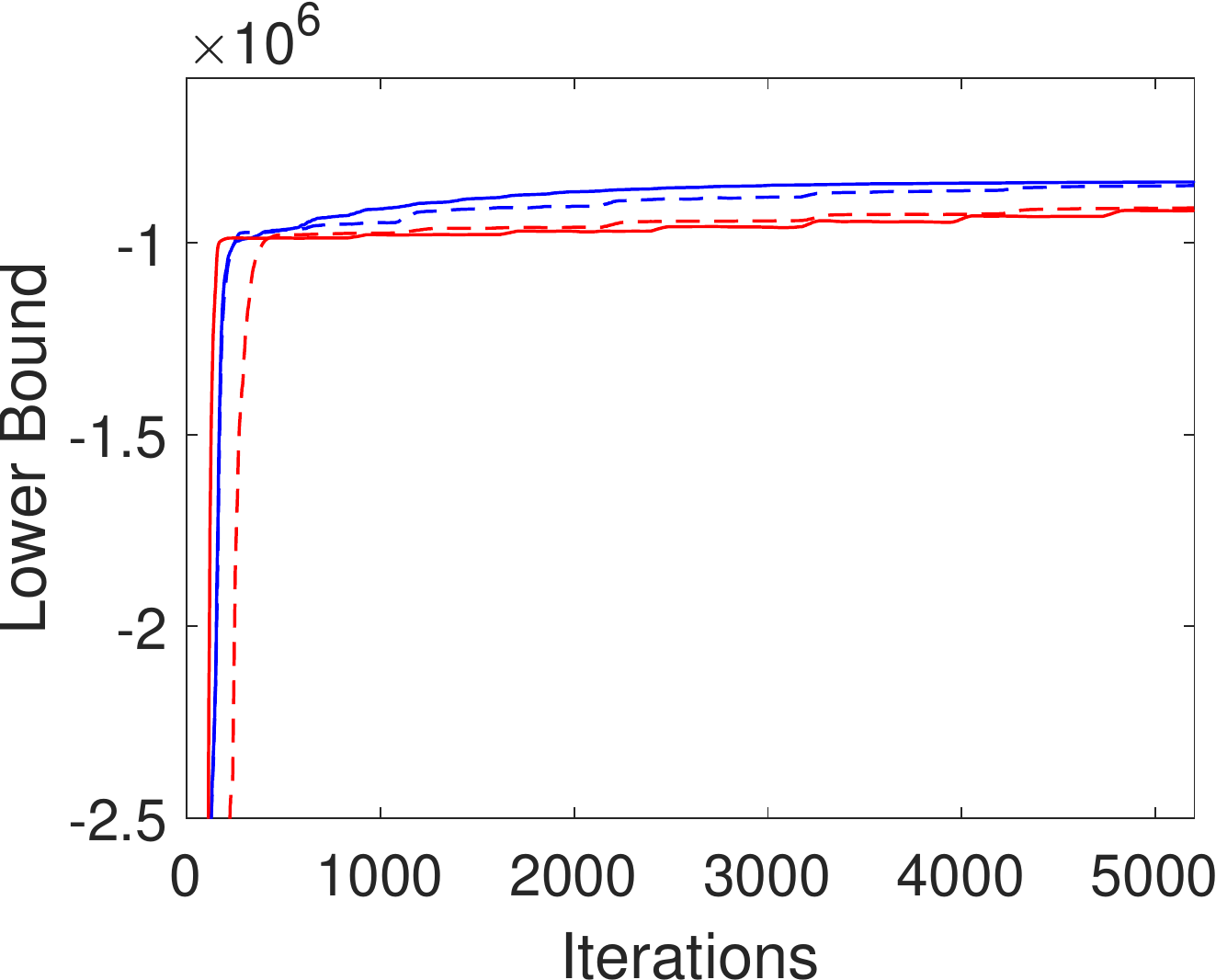}} &  
{\includegraphics[scale=0.3]  
{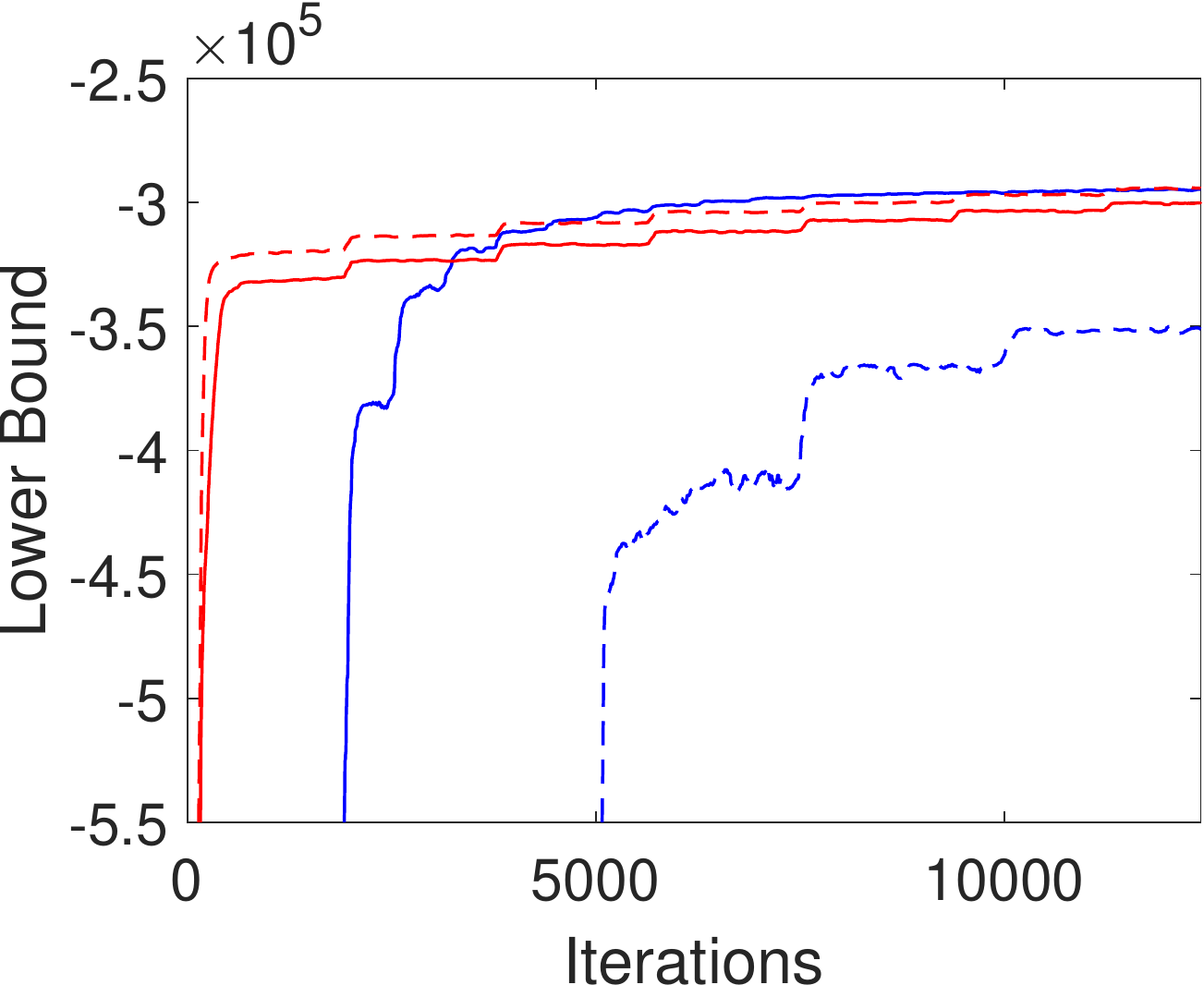}} \\
(a) & (b) & (c) \\
{\includegraphics[scale=0.3]  
{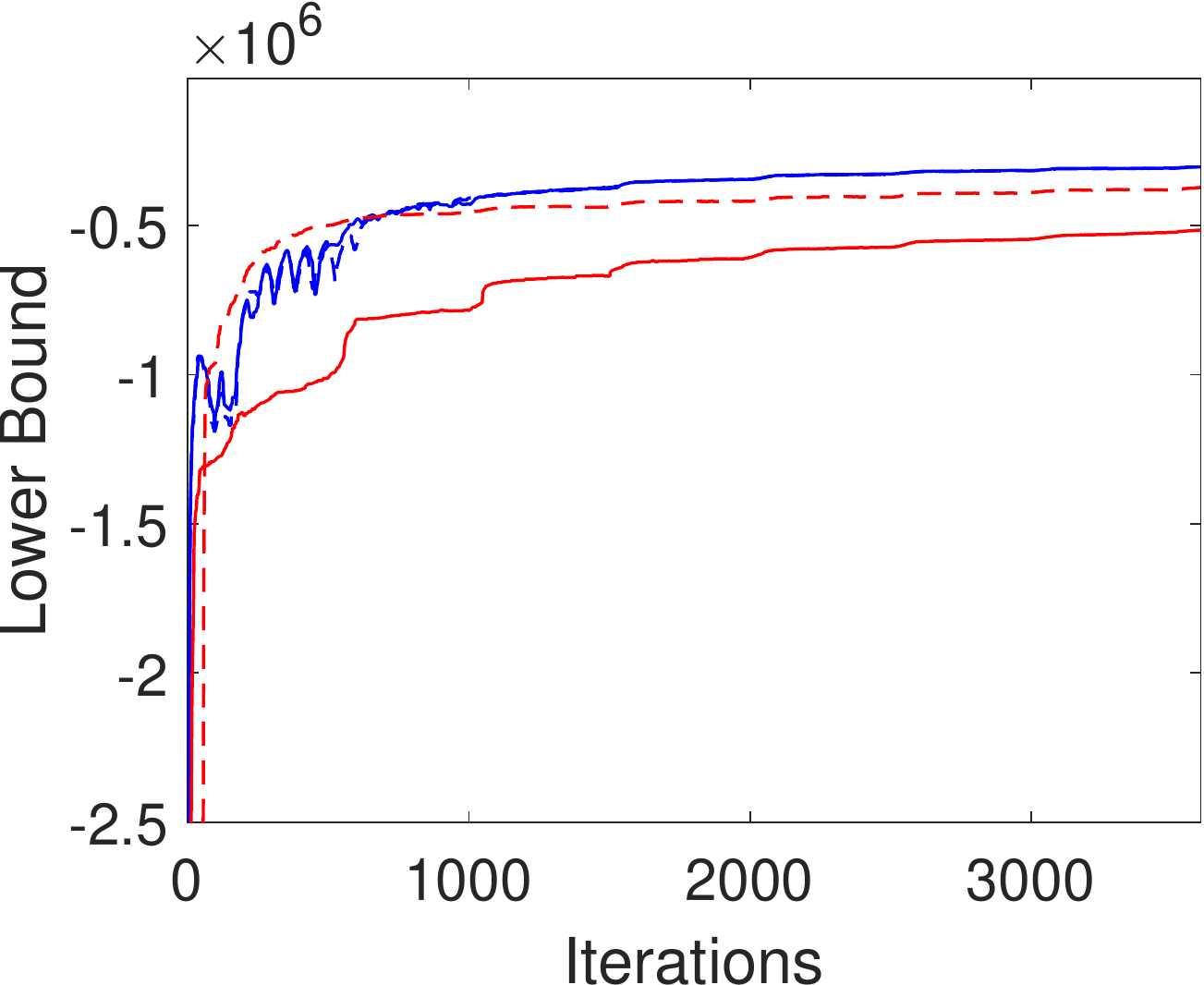}} & 
{\includegraphics[scale=0.3]  
{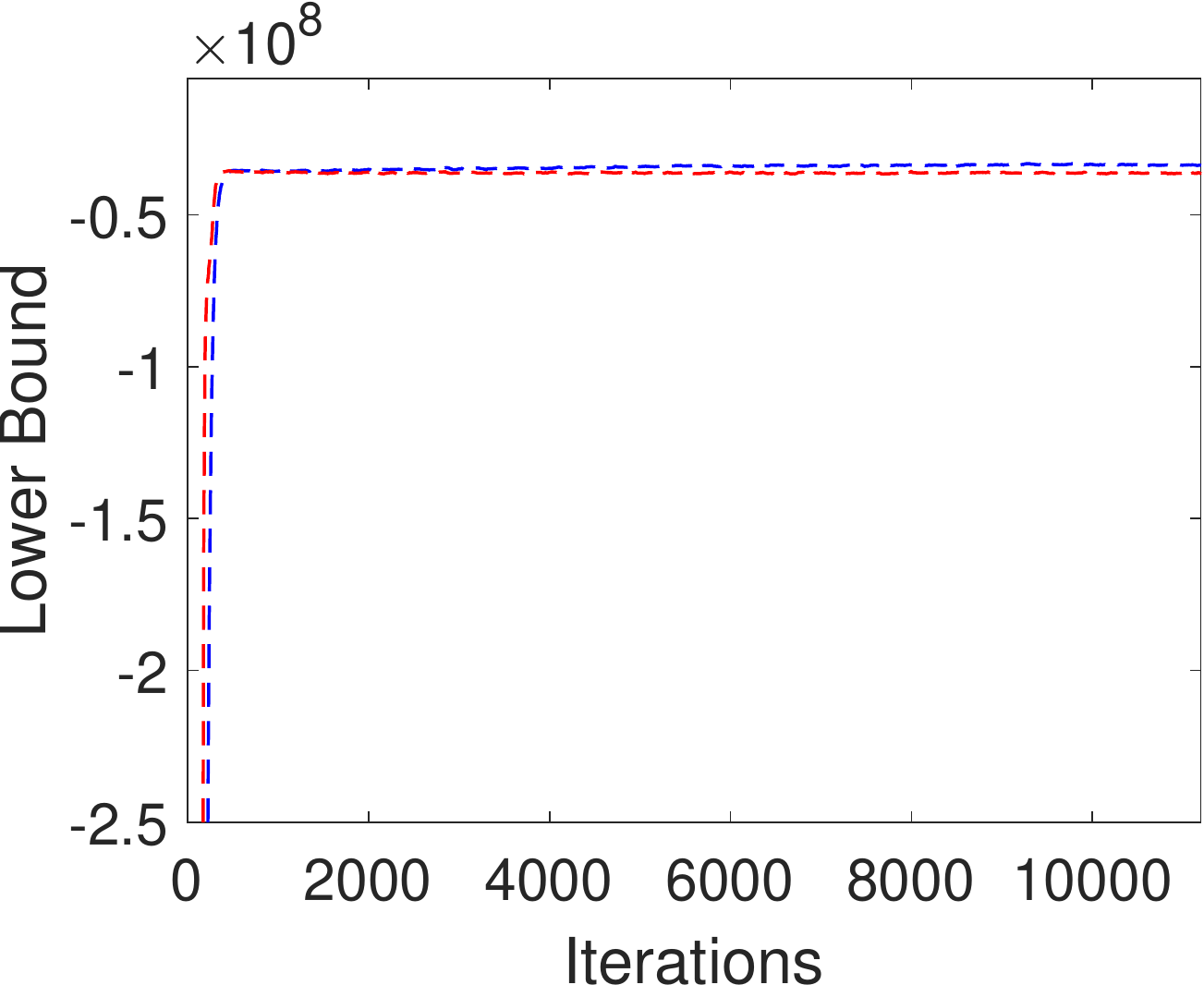}}   \\
(d) & (e) & 
\end{tabular}
\caption{Lower bounds of (a) Bibtex, (b) Delicious, (c) Mediamill, (d) EUR-Lex, and (e) AmazonCat. The solid lines correspond to the methods that kept fixed the inducing points while the dashed ones correspond to those that kept fixed the subspace inducing points.  Blue color suggests the use of linear kernel while the red one the use of squared exponential kernel.} 
\label{fig:lowerbounds_fixed}
\end{figure}

\begin{table}
\caption{Computational time (in minutes per epoch) of the MLGPF model for the seven multi-label datasets. 
}
\label{tbl_2_appendix}
\vskip 0.15in
\centering 
\begin{small}
\begin{tabular}{ccccc}
\toprule
Dataset     & s-{\sc linear}   & {\sc linear}      & s-{\sc se}   & {\sc se}   \\
\midrule
Bibtex      & 0.94             & 0.97              & 1.27         & 0.99       \\
Delicious   & 2.47             & 2.56              & 2.52         & 2.55       \\
Mediamill   & 5.96             & 5.70              & 6.64         & 6.0        \\
EUR-Lex     & 2.83             & 2.72              & 2.75         & 2.72       \\
RCV1        & 90.0             & 130.0             & 82.1         & 127.5      \\
AmazonCat   & 400.7            & 782.1             & 408.2        & 778.8      \\
\bottomrule
\end{tabular}
\end{small} 
\vskip -0.1in
\end{table}

\subsection{Extra experimental results \label{sec:extra_results}}
Table \ref{tbl_1_appendix} includes the predictive performance of the MLGPF model with fixed inducing inputs or fixed subspace inducing inputs. The case where our model is used with linear kernel and fixed subspace inducing inputs is denoted as sf-{\sc linear} while the case where a linear kernel is employed with fixed inducing inputs is denoted as f-{\sc linear} (similarly for the SE kernel). The experimental settings for each of the above methods are the same with the ones described in Section 4 of the main paper. Finally, in Figure \ref{fig:lowerbounds_fixed} can be found the evolution of the lower bound for each of the dataset in table \ref{tbl_1_appendix} while Figure \ref{fig:lowerbounds_opt} shows the corresponding lower bounds from the experiments of Section 4 of the main paper.

\end{document}